\documentclass{article}
\usepackage{iclr2027_conference,times}
\usepackage{microtype}

\usepackage{amsmath,amsfonts,bm}

\def\eqref#1{equation~\ref{#1}}

\def\1{\bm{1}}

\DeclareMathAlphabet{\mathsfit}{\encodingdefault}{\sfdefault}{m}{sl}
\SetMathAlphabet{\mathsfit}{bold}{\encodingdefault}{\sfdefault}{bx}{n}

\usepackage{url}
\usepackage[table]{xcolor}
\usepackage{wrapfig}
\usepackage{needspace}
\usepackage{booktabs}
\usepackage{graphicx}
\usepackage{listings}
\usepackage{multirow}
\usepackage{enumitem}
\usepackage{amsmath}
\usepackage{amssymb}
\usepackage{csquotes}
\usepackage{subcaption}
\usepackage{placeins}
\usepackage{fontawesome5}
\usepackage{etoolbox}
\usepackage{hyperref}
\hypersetup{
  hidelinks,
  pdftitle={Trajectory-Retrieval Speculative Decoding: When Does a Model's Own History Help?},
  pdfauthor={Yuyang Dai and Yushun Dong}
}

\definecolor{histmid}{RGB}{71,85,105}
\definecolor{promptgray}{HTML}{707780}
\definecolor{histblue}{HTML}{3F5F7F}
\definecolor{oursbg}{HTML}{EAF4EE}
\definecolor{zebra}{HTML}{F6F8FA}

\lstdefinestyle{prompt}{
  basicstyle=\ttfamily\footnotesize,
  breaklines=true,
  breakatwhitespace=false,
  columns=fullflexible,
  frame=single,
  framerule=0.3pt,
  xleftmargin=0.5em,
  xrightmargin=0.5em
}

\title{Trajectory-Retrieval Speculative Decoding:\\
When Does a Model's Own History Help?}

\author{
Yuyang Dai \qquad Yushun Dong \\
\normalfont Florida State University \\
\href{mailto:yd26@fsu.edu}{\texttt{yd26@fsu.edu}}
\qquad
\href{mailto:yd24f@fsu.edu}{\texttt{yd24f@fsu.edu}} \\[0.6em]
\href{https://github.com/YUYANG0915/tlr}{\faGithub\enspace\normalfont GitHub: \texttt{YUYANG0915/tlr}}
}

\iclrfinalcopy

\makeatletter
\patchcmd{\@maketitle}
  {\hsize\textwidth}
  {\hsize\textwidth\centering}
  {}
  {\PackageError{tlar-preprint}{Title centering patch failed}{Check the conference style definition of @maketitle.}}
\patchcmd{\@maketitle}
  {\begin{tabular}[t]{l}\bf\rule{\z@}{24pt}\@author\end{tabular}}
  {\begin{tabular}[t]{c}\bf\rule{\z@}{24pt}\@author\end{tabular}}
  {}
  {\PackageError{tlar-preprint}{Author centering patch failed}{Check the conference style definition of @maketitle.}}
\makeatother

\begin{document}
\maketitle
\fancyhead[L]{Preprint}

\begin{abstract}
Long chain-of-thought reasoning increases sequential decoding cost while creating a growing history of potentially reusable continuations. We investigate when this history supplies useful drafts and complements an existing drafter. Controlled source comparisons reveal trajectory-specific reuse, motivating our method \textbf{T}rajectory-\textbf{L}ocal \textbf{A}daptive \textbf{R}etrieval (TLAR). TLAR retrieves approximately matched continuations from the current trajectory and uses recent verification outcomes to adapt retrieval activation and candidate width. TLAR combines retrieved continuations with model-generated drafts in a shared candidate tree, preserving the target model's output distribution through exact verification. Across code debugging, mathematics, and open-ended writing, our evaluation connects source reuse, incremental acceptance, and execution cost. Combining TLAR with strong retrieval baselines improves token acceptance under matched verification budgets and increases end-to-end throughput over the draft-model baseline. These findings support generated trajectories as runtime memory for adaptive inference.
\end{abstract}

\begingroup
\setlength{\parskip}{2.5pt plus 0.5pt minus 0.3pt}
\setlength{\textfloatsep}{5pt plus 2pt minus 1pt}
\setlength{\floatsep}{4pt plus 2pt minus 1pt}
\setlength{\intextsep}{8pt plus 1pt minus 1pt}
\captionsetup{skip=4pt}
\captionsetup[subfigure]{skip=2pt}

\section{Introduction}
\label{intro}

Large language models (LLMs) can improve their performance on difficult tasks by generating long chains of thought (CoT) before producing a final answer~\citep{wei2022chain}. Long-CoT reasoning is used in competition mathematics~\citep{lightman2023lets,brown2024large,
muennighoff2025s1}, program synthesis and debugging~\citep{jimenez2024swebench,
jain2024livecodebench}, and scientific reasoning~\citep{rein2024gpqa}. This benefit
comes at a cost: a single response may contain thousands of reasoning tokens, as autoregressive models generate these tokens one at a time, each new token can be produced only after the preceding tokens are available~\citep{leviathan2023fast}. As reasoning traces grow longer, this sequential decoding cost accumulates, making efficient token generation increasingly important for practical long-CoT inference.

A practical goal is to accelerate this process without changing the target model's output distribution. Recently, speculative decoding does so by using a faster drafter to propose several future tokens, and the target model then verifies in parallel~\citep{leviathan2023fast,chen2023accelerating}. Most methods use a smaller model or a lightweight prediction head trained to imitate
the target model~\citep{li2024eagle,cai2024medusa}. Currently, some methods extend the
same idea from individual tokens to complete reasoning steps~\citep{
pan2026specreason,wang2026efficient,shi-etal-2025-speccot}. These methods can be
effective, but they require an additional trained component. To avoid this requirement, retrieval-based methods construct drafts from existing text. They search the prompt, an external datastore, or tokens already generated by the target model, and use the continuations following a match as draft candidates~\citep{somasundaram2025pld+,he2024rest,fu2024lookahead,
hu2025sam,oliaro2026suffixdecoding,li2024nearest,luo2025turning,chen2025rapid}.
Among these, we focus on the tokens generated earlier in the current trajectory. Similar contexts may appear more than once during long reasoning. When this happens, the tokens that followed an earlier context can be reused as a draft. However, using the current trajectory as a draft source raises three challenges.

\noindent\textbf{Challenge 1: Separating trajectory reuse from common text.}
Retrieval may succeed because reasoning traces share common text. Different reasoning traces may share common words, symbols, and formatting patterns, allowing retrieval to succeed even when the current history is not used. High acceptance can also result from a model repeating the same text without making progress. A fair test must compare the current trajectory with other retrieval sources while keeping the task, target model, and matching rule unchanged.

\noindent\textbf{Challenge 2: Finding tokens beyond a small draft model.}
Even if the current trajectory provides useful drafts, a small draft model may already predict the same tokens. In that case, adding retrieval does not
increase the number of accepted tokens. So we must compare the combined system with the same small draft model used alone.

\noindent\textbf{Challenge 3: Controlling the cost of retrieval verification.}
Every retrieved candidate must be verified by the target model which adds computation, while reusable content does not appear at every position in a
reasoning trace. An effective method should decide when to use retrieval and how many candidates to verify based only on results observed at earlier positions.

We address these challenges through source comparisons and
\textbf{T}rajectory-\textbf{L}ocal \textbf{A}daptive \textbf{R}etrieval (TLAR). After defining acceptance
and cost in Section~\ref{prelim}, we compare current history with control
sources and another rollout of the same problem in Section~\ref{sec:exploratory}.
These observations guide TLAR in Section~\ref{sec:method}: we find approximate
context matches in generated history and combine their continuations with
SmallDraft in a prefix-sharing tree. We retain the SmallDraft candidate and
use earlier verification results to choose when to retrieve and how many
candidates to include. Periodic probes check for new useful content, and the
target model verifies every proposed token exactly. In
Section~\ref{sec:results}, we test added acceptance over SmallDraft, added
coverage under shared retrieval-node budgets, and the controller's gain--cost trade-off
separately.
We summarize our contributions as follows:


\begin{itemize}
    \item \textbf{Uncovering Trajectory-Specific Reuse.}
    We test when the current history provides more accepted tokens than other sources, and use same-prompt controls to distinguish the effect of the trajectory from that of the problem. The advantage depends on model and task.

    \item \textbf{More Accepted Tokens Within the Same Budget.}
    We measure how many additional tokens the target model accepts when TLAR is combined with a draft model compared with using the same draft model alone. We then combine TLAR with the strongest retrieval baseline under shared verification-node budgets. The union yields more accepted tokens than either source alone, showing the value of trajectory-local candidates within the budget.
    
    \item \textbf{An Adaptive Retrieval Method.}
    We introduce TLAR, which searches the current trajectory and uses earlier verification results to decide when to retrieve and how many candidates to submit. Experiments across models, tasks, and verification budgets show when this approach is useful and how its gain changes with its cost.
\end{itemize}

\vspace{-0.5mm}
\section{Preliminaries}
\label{prelim}

\paragraph{Speculative Decoding and Token Verification.}
Given a prompt \(x\), a target language model \(\mathcal{M}\) generates a token sequence \(\mathbf{y}=(y_1,\ldots,y_T)\). 
At position \(t\), the tokens already
generated form the history \(h_t=(y_1,\ldots,y_{t-1})\). In speculative decoding, a drafter uses this history to propose a sequence of future tokens
\(\mathbf{c}=(c_1,\ldots,c_L)\). The target model verifies these tokens in one forward pass and accepts the longest prefix that matches
\((y_t,\ldots,y_{t+L-1})\). We define the accepted length as \(a(\mathbf{c},t)=\max\{\ell:0\leq\ell\leq L,\, c_{1:\ell}=y_{t:t+\ell-1}\}\). 
Then we first generate one target sequence \(\mathbf{y}\) for each prompt, and test every drafting method at the same positions in this
sequence, so all methods are asked to predict the same next tokens.

\paragraph{Model-Based and Retrieval-Based Drafting.}
The candidate \(\mathbf{c}\) defined above can be produced in two ways. A model-based drafter uses a smaller language model \(\mathcal{M}_{\mathrm{draft}}\) to predict the next tokens from the prompt and
the current history~\citep{li2024eagle,cai2024medusa}. Let \(\mathbf{d}_t\) denote its draft at position \(t\). Its accepted length is \(a_{\mathrm{small}}(t)=a(\mathbf{d}_t,t)\).
A retrieval-based drafter constructs candidates from text that is already available~\citep{somasundaram2025pld+,he2024rest,hu2025sam, oliaro2026suffixdecoding}. It searches a text source for a context similar to the current one and uses the tokens following that context as a draft. In this work, we focus on retrieval from the current history \(h_t\). Let \(\mathcal{R}_t\) denote the set of candidates retrieved at position \(t\). Its
accepted length is \(a_{\mathrm{retr}}(t)=\max_{\mathbf{r}\in\mathcal{R}_t}
a(\mathbf{r},t)\), with \(a_{\mathrm{retr}}(t)=0\) when no candidate is retrieved.

\noindent \paragraph{Incremental Acceptance and Verification Cost.}

We now measure whether retrieval adds value beyond the small draft model. We first consider positions where the small draft model fails on its first token, so \(a_{\mathrm{small}}(t)=0\). The recovery rate is
\(R=\Pr(a_{\mathrm{retr}}(t)\geq 1\mid a_{\mathrm{small}}(t)=0)\). It measures how often retrieval finds at least one correct token when the small draft model fails immediately.
Since recovery does not cover positions where the small draft model predicts some correct tokens but retrieval provides a longer continuation, we define the added accepted tokens at position \(t\) as
\(\delta(t)=\max\{a_{\mathrm{small}}(t),a_{\mathrm{retr}}(t)\}
-a_{\mathrm{small}}(t)\). The average gain is
\(\Delta=\frac{1}{T}\sum_{t=1}^{T}\delta(t)\). Retrieval receives credit only when it allows the target model to accept more tokens than the same small draft model used alone. We then measure verification cost by the number of retrieval nodes submitted to the
target model. Let \(N_{\mathrm{retr}}(t)\) denote this number at position \(t\), with \(N_{\mathrm{retr}}(t)=0\) when retrieval is not used. The average cost is
\(C=\frac{1}{T}\sum_{t=1}^{T}N_{\mathrm{retr}}(t)\).
\setlength{\parskip}{3pt plus 1pt minus 0.3pt}
\section{An Exploratory Study}
\label{sec:exploratory}

Before testing whether retrieval can improve a small draft model, we first ask whether the current trajectory is a better retrieval source than other available text. We then ask whether any advantage remains when the alternative trajectory comes from the same problem, and whether successful retrieval is spread evenly through generation.

\noindent \textbf{Retrieval and controller settings.}
The default retrieval configuration uses context length $c=4$, mismatch tolerance $\epsilon=1$, history threshold $\tau=512$, and continuation depth $d=4$. Fixed-width experiments use $k=4$. The adaptive controller uses $k_{\min}=1$, $k_{\max}=4$, $\rho_{\min}=0.05$, half-life $H=32$, and probe interval $P=32$. Section~\ref{sec:gain_cost} examines candidate configurations and controller components; Appendix~\ref{app:w7} reports the half-life and probe-interval sweeps.

\noindent \textbf{Evaluation Protocol.}
We keep the target trajectory and acceptance rule fixed and compare four draft
sources. \emph{Within} searches the current generated history. \emph{Cross}
searches another trajectory from the same target model and dataset, restricted
to the same available history length. \emph{Shuffled} searches a block-shuffled
version of the current history. \emph{Unigram} uses token frequencies without
matching the current context. Within, Cross, and Shuffled use the same matcher.
We report the mean exactly accepted length, \(E_{\mathrm{exact}}\), for
Qwen3-8B, Mistral-Small-3.2-24B-Instruct, and Llama-3.1-8B-Instruct.

\begin{figure}[t]
\centering
\begin{subfigure}[t]{0.32\linewidth}
\includegraphics[width=\linewidth]{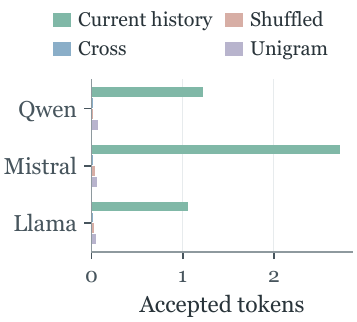}
\caption{Code debugging (100 tasks)}
\end{subfigure}\hfill
\begin{subfigure}[t]{0.32\linewidth}
\includegraphics[width=\linewidth]{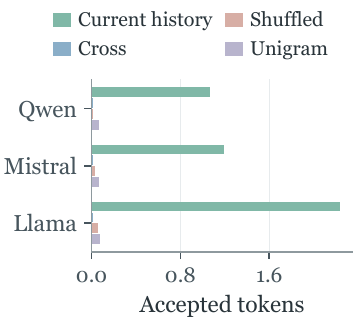}
\caption{Mathematics (24 tasks)}
\end{subfigure}\hfill
\begin{subfigure}[t]{0.32\linewidth}
\includegraphics[width=\linewidth]{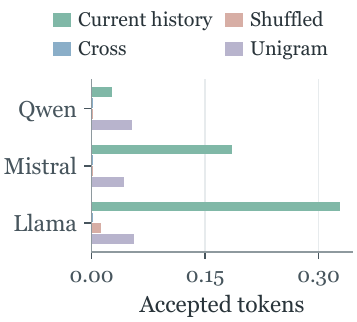}
\caption{Open-ended writing (24 tasks)}
\end{subfigure}
\caption{Current history outperforms the source controls in code and mathematics,
but not consistently in open-ended writing.
We compare mean exactly accepted tokens, \(E_{\mathrm{exact}}\), from four sources.
Current history (Within) uses the generated prefix. Cross uses another trajectory
from the same model and dataset, limited to the same history length.
Shuffled uses the current history with its blocks reordered. Unigram uses token
frequencies without matching the current context. Panels use separate horizontal
scales; Llama mathematics excludes repeated-copying cases (Appendix~\ref{app:selfcopy}).
Exact values appear in Appendix Table~\ref{tab:source_full}.}
\label{fig:selfcopy3x3}
\end{figure}

\noindent \textbf{Finding 1: Current history helps more in code and mathematics.}
We find higher accepted lengths from Within than from each of the three controls
for all three models in code debugging and mathematics
(Figure~\ref{fig:selfcopy3x3}).
Changing the source changes acceptance even when the target tokens
remain fixed. The effect is smaller in open-ended writing: Mistral's Within
advantage becomes smaller, and Qwen does not exceed Unigram. We also retain the
Llama repeated-copying audit, since long matches caused by repeated text do not
by themselves show useful reuse. These differences limit the claim to the
models and tasks in which the source controls support it.

\begin{figure}[t]
\centering
\begin{subfigure}[t]{0.32\linewidth}
\includegraphics[width=\linewidth]{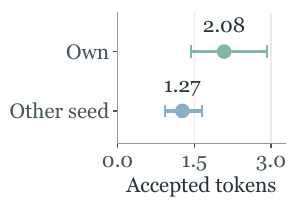}
\caption{Qwen3-8B: exact context}
\end{subfigure}\hfill
\begin{subfigure}[t]{0.32\linewidth}
\includegraphics[width=\linewidth]{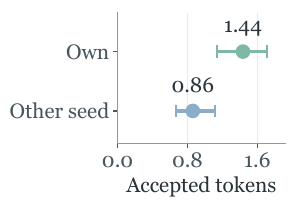}
\caption{Mistral-24B: exact context}
\end{subfigure}\hfill
\begin{subfigure}[t]{0.32\linewidth}
\includegraphics[width=\linewidth]{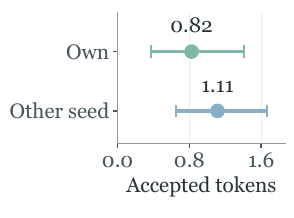}
\caption{Llama-3.1-8B: exact context}
\end{subfigure}
\caption{Own-history and same-prompt, different-seed retrieval. Points show mean accepted tokens and bars show 95\% bootstrap intervals. Each comparison uses 24 paired problems.}
\label{fig:g3c}
\end{figure}

\noindent \textbf{Finding 2: The same problem need not provide the same drafts.}
\label{sec:trajectory_control}
Cross changes both the problem and the trajectory. To separate them, we compare two rollouts of each of 24 code-debugging problems with different random seeds.
We show the paired source comparisons in Figure~\ref{fig:g3c}, comparing own and other history within each panel. Qwen and Mistral favor their own history over the other rollout, while Llama shows the opposite ordering. These paired comparisons support trajectory-specific reuse for Qwen and Mistral, whereas Llama's reusable drafts appear to depend more on the shared problem than on its own rollout.

\noindent \textbf{Finding 3: Retrieval success is concentrated at nearby positions.}
Finally, successful retrieval is not evenly distributed. In the code-debugging retrieval measurements, a hit is more likely after another hit than at an arbitrary position for all three models. Thus, recent verification results provide information about nearby positions. This motivates the running score and periodic probes in Section~\ref{sec:method}. Whether using this information saves enough verification work is a separate question, we tested in Section~\ref{sec:gain_cost}. Together, these findings motivate searching current history and adapting
retrieval during generation. We combine the two
sources under target verification in Section~\ref{sec:method}, then measure
their added acceptance and cost separately in Section~\ref{sec:results}.

\vspace{-0.5mm}
\section{Method: \textbf{T}rajectory-\textbf{L}ocal \textbf{A}daptive \textbf{R}etrieval}
\label{sec:method}

\begin{figure}[t]
\centering
\includegraphics[width=\linewidth]{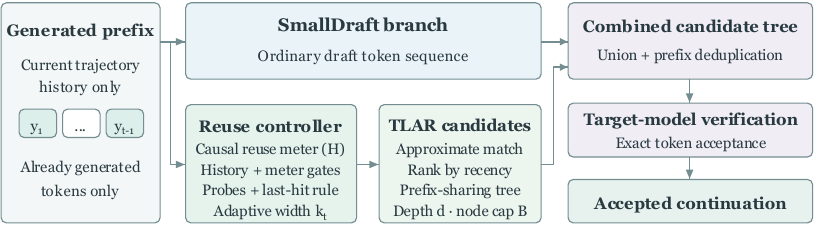}
\caption{Trajectory-Local Adaptive Retrieval.
TLAR uses earlier verification results to decide whether retrieval should run and how many candidates to retrieve from the current history. The retrieved candidates are combined with the SmallDraft candidate and verified by the
target model.}
\label{fig:method_pipeline}
\end{figure}

TLAR adds a controlled history search to SmallDraft, as we show in Figure~\ref{fig:method_pipeline}. We describe the method in four parts. First, we find approximate context matches in generated history, rank them by recency, and place their continuations in a prefix-sharing tree (Section~\ref{method:retrieval}). Next, we track earlier verified retrieval results in a running score (Section~\ref{method:meter}). We use this score,
the last retrieval result, and periodic probes to choose when to retrieve and how many candidates to keep (Section~\ref{method:controller}). Finally, we
combine these candidates with SmallDraft, remove duplicate prefixes, and verify the resulting tree with the target model (Section~\ref{method:hybrid}).

\vspace{-0.5mm}
\subsection{Retrieving Candidates from the Current History}
\label{method:retrieval}

At position \(t\), TLAR searches only the generated history \(h_t=(y_1,\ldots,y_{t-1})\). This choice follows from Section~\ref{sec:exploratory}, where the current history provides better candidates than the control sources in the settings where retrieval is useful. Let \(\mathbf{u}_t=(y_{t-c},\ldots,y_{t-1})\) be the most recent \(c\) tokens.
For an earlier position \(j\), let \(\mathbf{u}_j=(y_{j-c},\ldots,y_{j-1})\) be the context that appeared before \(y_j\). TLAR identifies earlier positions whose contexts match the current context within a Hamming distance of $\epsilon$ and whose next $d$ tokens are fully contained in the generated history:
\begin{equation}
\mathcal{J}_t =
\left\{
j : c+1 \leq j \leq t-d,\;
d_{\mathrm{Ham}}(\mathbf{u}_t,\mathbf{u}_j) \leq \epsilon
\right\}.
\label{eq:historical_offsets}
\end{equation}

where \(d_{\mathrm{Ham}}\) counts the token differences between two contexts. The value \(\epsilon\) is the largest number of differences allowed. The condition \(j\leq t-d\) ensures that the continuation \((y_j,\ldots,y_{j+d-1})\) is already contained in \(h_t\). TLAR cannot use current or future target tokens when constructing a candidate.
For every \(j\in\mathcal{J}_t\), TLAR takes the next \(d\) historical tokens \(\mathbf{r}_j=(y_j,\ldots,y_{j+d-1})\) as a candidate. Allowing up to \(\epsilon\) context differences prevents small changes in punctuation, formatting, or wording from removing an otherwise useful candidate. Approximate matching is used only to find candidate locations. The retrieved tokens are still accepted using the exact verification rule defined in
Section~\ref{prelim}. TLAR ranks the matching positions by recency and selects the first \(k_t\). Their continuations are inserted into a prefix-sharing tree. When two
candidates begin with the same tokens, their shared prefix appears only once in the tree. The target model does not need to verify duplicate prefixes. Section~\ref{method:controller} explains how TLAR chooses \(k_t\).

\vspace{-0.5mm}
\subsection{Tracking Recent Retrieval Success}
\label{method:meter}

Let \(m_t\in\{0,1\}\) indicate whether retrieval is attempted at position \(t\).
When \(m_t=1\), let \(z_t=\mathbb{I}[a_{\mathrm{retr}}(t)\geq 1]\) indicate whether the target model
accepts at least one retrieved token. 
TLAR stores these results in a running score \(\hat{\rho}_t\), initialized as \(\hat{\rho}_1=0\) and updated after verification:

\begin{equation}
\hat{\rho}_{t+1}
=
\begin{cases}
(1-\alpha)\hat{\rho}_t+\alpha z_t, & m_t=1,\\
(1-\alpha)\hat{\rho}_t, & m_t=0.
\end{cases}
\label{eq:reuse_meter}
\end{equation}

When retrieval is attempted, the first case adds the latest result to the score. When retrieval is not attempted, the second case gradually reduces the
effect of older results. A larger \(\hat{\rho}_t\) means that retrieval has worked more often at recent positions. The score uses only verification results observed before position \(t\). It does not require another trained predictor, an external trajectory, or
additional target-model outputs. We set \(\alpha=1-2^{-1/H}\), where \(H\) controls how quickly earlier results lose weight. We use \(H=32\) by default and test other values in Section~\ref{sec:gain_cost}.

\vspace{-0.5mm}
\subsection{Choosing When and How Much to Retrieve}
\label{method:controller}

TLAR uses recent verification feedback to allocate retrieval work along the generated trajectory. The controller first requires $|h_t|>\tau$, allowing historical contexts and their continuations to accumulate. Once the history passes this threshold, retrieval activation combines a running success score, the latest observed
retrieval outcome, and periodic probes. These signals capture sustained reuse, respond to newly observed success, and discover reuse as generation enters new regions.
Let $q_t\in\{0,1\}$ indicate a probe position, and let $s(t)<t$ denote the most recent retrieval attempt, with $z_{s(t)}=0$ before the first attempt. The activation rule is $g_t=\mathbb{I}[|h_t|>\tau]\,
\mathbb{I}[\hat{\rho}_t\geq\rho_{\min}\lor z_{s(t)}=1\lor q_t=1]$.
The running score sustains retrieval across regions with repeated success. The latest outcome allows a successful attempt to trigger retrieval at the next decision even while the running score remains below the threshold. During inactive periods, the score decays and periodic probes collect fresh verification feedback. A successful probe can reactivate retrieval as useful continuations become available.

For each active decision, TLAR also allocates a candidate width. At an active non-probe position, it sets
$k_t=\operatorname{clip}
(\lceil k_{\max}\hat{\rho}_t\rceil,k_{\min},k_{\max})$;
at a probe position, it sets $k_t=k_{\min}$.
This rule ties the search breadth to accumulated evidence of
reuse: lower scores admit fewer historical continuations,
while higher scores expand candidate coverage.
The selected continuations enter a prefix-sharing tree, so
their verification cost also depends on how many prefixes
they share. Candidate width controls the proposed work, and
the distinct-node count measures the resulting tree size.
Activation and width form a feedback loop with target verification. Activation determines when to collect retrieval evidence; width determines how many continuations contribute candidates at that decision. After verification, the observed retrieval outcome updates the score used for subsequent decisions.
Section~\ref{sec:gain_cost} evaluates these components through activation and width ablations, probe removal, and parameter sensitivity experiments.

\vspace{-0.5mm}
\subsection{Combining TLAR with SmallDraft}
\label{method:hybrid}

At each position, SmallDraft produces the candidate
\(\mathbf{d}_t\) defined in Section~\ref{prelim}. If \(g_t=1\), TLAR also constructs a candidate tree from the current history. If \(g_t=0\), the TLAR
candidate set is empty.
The SmallDraft candidate and the TLAR candidates are inserted into one prefix-sharing tree. Duplicate prefixes are merged before verification, and
the target model verifies the resulting tree. After verification, TLAR uses the retrieval result to update \(\hat{\rho}_{t+1}\).
We evaluate the combined system using the recovery rate \(R\), added accepted tokens \(\Delta\), and verification cost \(C\) defined in Section~\ref{prelim}. When TLAR is inactive, we set
\(a_{\mathrm{retr}}(t)=0\) and \(N_{\mathrm{retr}}(t)=0\). The evaluation measures the complete method, including positions where its controller decides not to retrieve.

\section{Experiments: Utility, Specificity, and Complementarity}
\label{sec:results}

Section~\ref{sec:exploratory} tests where reusable tokens come from. We now ask
whether they add value beyond SmallDraft (\textbf{RQ1}), whether they improve
coverage alongside another retrieval source under the same verification budget
(\textbf{RQ2}), and how the controller changes gain and
verification cost (\textbf{RQ3}). These questions distinguish a useful source
from a useful addition to an existing drafter.

\subsection{Model and Evaluation Settings}
\label{sec:setup}

We evaluate Qwen3-8B, Mistral-24B, and Llama-3.1-8B-Instruct on 100 code-debugging,
24 mathematical-reasoning, and 24 open-ended-writing problems. SmallDraft is
Qwen3-0.6B, Ministral-8B-Instruct-2410, and Llama-3.2-1B-Instruct, respectively.
For each target, the offline comparisons use the same SmallDraft predictions
and fixed target trajectories. Code debugging is the main setting; the other
domains test whether the observed gains extend beyond it.
Keeping each target trajectory fixed lets us compare candidate acceptance
without changing the target tokens between methods. We do not compare one method's generated response with a different response from another method.
The source controls and same-prompt tests in Section~\ref{sec:exploratory} answer a different question and are not counted as additional tests of
SmallDraft complementarity.

We report additional exactly accepted tokens per position, \(\Delta\), over
SmallDraft alone. We first fix the history gate and candidate width, then test
the adaptive controller separately. For RQ2, we compare TLAR, the strongest
single-source retrieval baseline, and their budgeted union under shared caps
of \(B\in\{4,8\}\) distinct retrieval nodes per position. Union candidates
are merged into one prefix-sharing tree, deduplicated, and pruned to the common
node budget.
Acceptance comes from offline replay, and node counts describe retrieval-side work. We report 95\% bootstrap intervals where available. Appendix~\ref{app:revision_results} records the settings and cost measures,
while Appendix~\ref{app:causal_validation} reports the completed causal-boundary
and event-driven replay audits. Appendix~\ref{app:wallclock} separately evaluates
the complete adaptive union tree under matched end-to-end serving conditions. We keep the three comparisons separate: RQ1 adds retrieval to SmallDraft,
RQ2 compares single-source retrieval with a multi-source union under the same
verification-node budget, and RQ3 changes TLAR's controller.
Each uses its own matched reference. In particular, a larger accepted-token increment does not by itself show that its verification work is worthwhile.

\begin{figure}[t]
\centering
\textbf{Accepted tokens from history: source test}\par\smallskip
\begin{subfigure}[t]{0.32\linewidth}
\includegraphics[width=\linewidth]{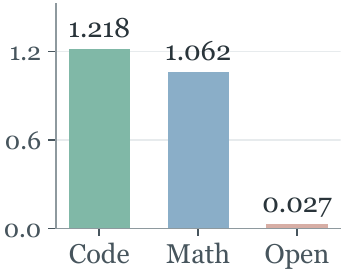}
\caption{Qwen3-8B}
\end{subfigure}\hfill
\begin{subfigure}[t]{0.32\linewidth}
\includegraphics[width=\linewidth]{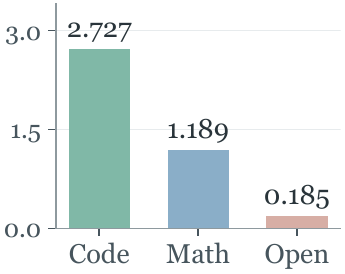}
\caption{Mistral-24B}
\end{subfigure}\hfill
\begin{subfigure}[t]{0.32\linewidth}
\includegraphics[width=\linewidth]{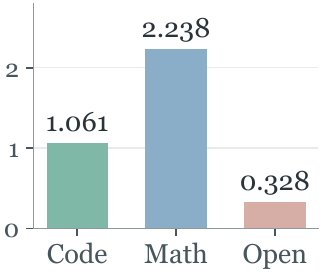}
\caption{Llama-3.1-8B}
\end{subfigure}
\par\medskip
\textbf{Extra tokens from fixed-width TLAR over SmallDraft}\par\smallskip
\begin{subfigure}[t]{0.32\linewidth}
\includegraphics[width=\linewidth]{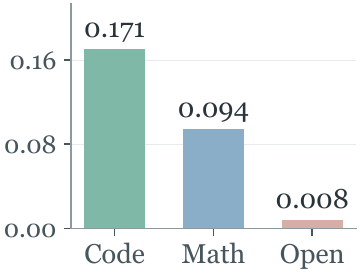}
\caption{Qwen3-8B}
\end{subfigure}\hfill
\begin{subfigure}[t]{0.32\linewidth}
\includegraphics[width=\linewidth]{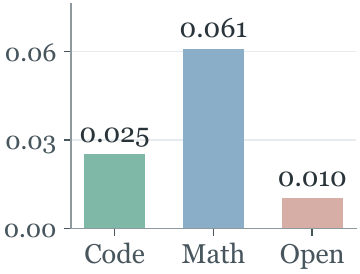}
\caption{Mistral-24B}
\end{subfigure}\hfill
\begin{subfigure}[t]{0.32\linewidth}
\includegraphics[width=\linewidth]{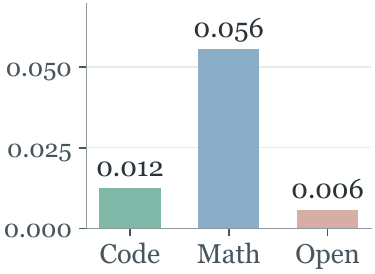}
\caption{Llama-3.1-8B}
\end{subfigure}
\caption{More reuse need not mean more added tokens.
For Mistral (middle column), history supplies more accepted tokens in code,
but retrieval adds more beyond SmallDraft in mathematics.
Top: Within-source \(E_{\mathrm{exact}}\) from Section~\ref{sec:exploratory}.
Bottom: fixed-gate, fixed-width TLAR's increment over the same SmallDraft,
\(\Delta=T^{-1}\sum_t[\max(a_{\mathrm{small}}(t),a_{\mathrm{retr}}(t))-a_{\mathrm{small}}(t)]\)
on fixed target tokens. Code uses 100 tasks; Math and Open use 24 each. Each panel has its own scale.}
\label{fig:g3a}
\end{figure}

\subsection{RQ1: Does Trajectory-Local Memory Complement SmallDraft?}
\label{sec:main_results}

We first test whether retrieval supplies tokens that the same SmallDraft
misses. We use 100 code-debugging tasks per model, a fixed 512-token history
gate, and \(k=d=4\), without adaptive activation or width. We report added
accepted length in Table~\ref{tab:main_hybrid}; SmallDraft's own accepted length
\(e\) is shown separately. We retain the full node-cost comparison in
Appendix~\ref{app:main_cost}. In Figure~\ref{fig:g3a}, we place the source
measurements above TLAR's increment over SmallDraft for the same model and
task. The top row asks how much the history supplies in the source test.
The bottom row asks how much retrieval adds after SmallDraft's accepted
tokens are already counted and its reference is SmallDraft alone.

\setlength{\parskip}{2.5pt plus 0.5pt minus 0.3pt}
\begin{wraptable}{R}{0.57\linewidth}
\vspace{-8pt}
\centering
\small
\setlength{\tabcolsep}{3pt}
\captionsetup{justification=justified,singlelinecheck=false,margin=0pt}
\caption{Added acceptance over SmallDraft on code debugging.
Bold marks the union, which has the largest increment per model.}
\label{tab:main_hybrid}
\begin{tabular*}{\linewidth}{@{}l@{\extracolsep{\fill}}rrr@{}}
\toprule
Method & Qwen & Mistral & Llama \\
\midrule
SmallDraft ($e$) & 2.383 & 3.148 & 2.823 \\
\midrule
SuffixDecoding ($\Delta$) & 0.128 & 0.026 & 0.013 \\
STAND ($\Delta$) & 0.164 & 0.017 & 0.011 \\
TLAR (fixed $k$; $\Delta$) & 0.171 & 0.025 & 0.018 \\
\midrule
STAND$\cup$TLAR ($\Delta$)
& \textbf{0.228} & \textbf{0.032} & \textbf{0.026} \\
\bottomrule
\end{tabular*}
\vspace{-4pt}
\end{wraptable}
TLAR adds accepted tokens for all three targets. Its single-method point
estimate is highest for Qwen and Llama; SuffixDecoding is slightly higher for
Mistral. The uncapped union in Table~\ref{tab:main_hybrid} has the largest
accepted-token increment, motivating the matched-budget complementarity test
in Section~\ref{sec:rq2}. The domain comparison shows why source reuse and added
acceptance must be tested separately: Mistral gains more over SmallDraft in
mathematics than in code, despite stronger source reuse in code. Open-ended
increments remain small for all three targets. A useful history can overlap with what SmallDraft already predicts. The middle column makes the distinction concrete: Mistral has more accepted history tokens in code than in mathematics, but the added tokens over
SmallDraft follow the opposite ordering. Thus, choosing where to use retrieval
from source acceptance alone can be misleading.
We use the matched SmallDraft comparison to measure that added value directly.

\subsection{RQ2: Does TLAR Improve Coverage Under a Matched Verification Budget?}
\label{sec:rq2}
\label{sec:baseline_plugin}

We next test whether trajectory-local candidates improve coverage when they compete with another retrieval source for the same verification budget.
For each target, we compare TLAR, the strongest single-source baseline, and their budgeted union on code debugging. The strongest baseline is STAND for
Qwen and Mistral and SuffixDecoding for Llama. All methods receive the same cap of \(B\in\{4,8\}\) distinct retrieval nodes per position.
For the union, candidates from both sources enter one prefix-sharing tree.
Duplicate prefixes are merged, and the tree is pruned to the shared node budget. Ranking uses source-local recency and verification outcomes observed at earlier decoding events. Candidates from both sources compete for the
available tree nodes.

Figure~\ref{fig:matched_budget} shows that the budgeted union achieves the
largest accepted-token gain for all three targets at both budgets.
At \(B=4\), its gains over the strongest single-source baseline are
\(13.7\%\), \(11.5\%\), and \(5.0\%\) for Qwen, Mistral, and Llama,
respectively. At \(B=8\), the corresponding gains are \(16.8\%\),
\(14.8\%\), and \(4.8\%\). Trajectory-local candidates improve the
composition of the verification tree within the shared node cap.
Appendix Table~\ref{tab:matched_budget} provides the exact point estimates.

The relative improvement is largest for Qwen, followed by Mistral and Llama.
This pattern is consistent with the same-prompt source comparison in
Section~\ref{sec:exploratory}: Qwen and Mistral favor their own trajectory,
while Llama favors the other rollout. The source comparison and the budgeted
union test connect trajectory-specific reuse to added acceptance in the
hybrid drafter.

\begin{figure}[t]
\centering
\begin{subfigure}[t]{0.32\linewidth}
\includegraphics[width=\linewidth]{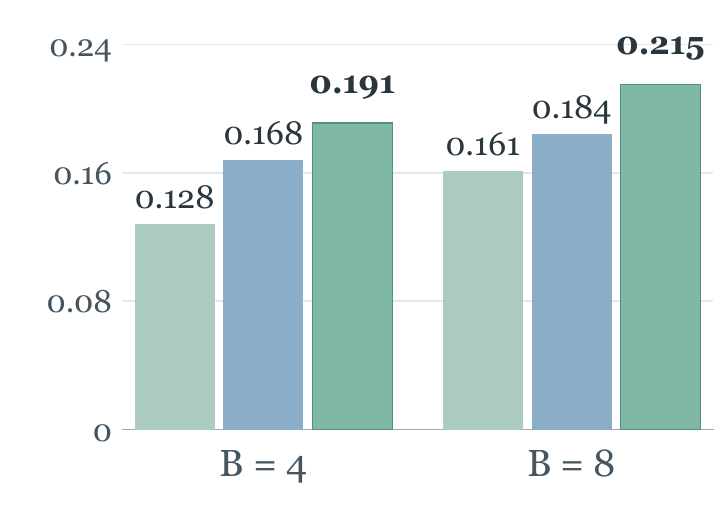}
\caption{Qwen3-8B}
\label{fig:matched_budget_qwen}
\end{subfigure}\hfill
\begin{subfigure}[t]{0.32\linewidth}
\includegraphics[width=\linewidth]{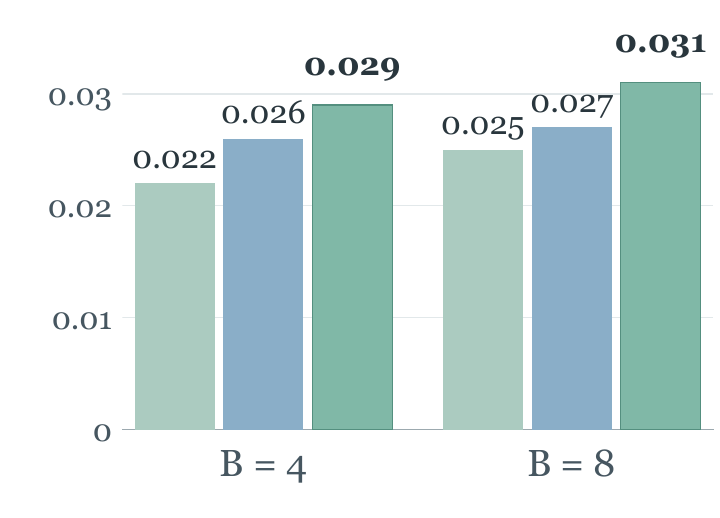}
\caption{Mistral-24B}
\label{fig:matched_budget_mistral}
\end{subfigure}\hfill
\begin{subfigure}[t]{0.32\linewidth}
\includegraphics[width=\linewidth]{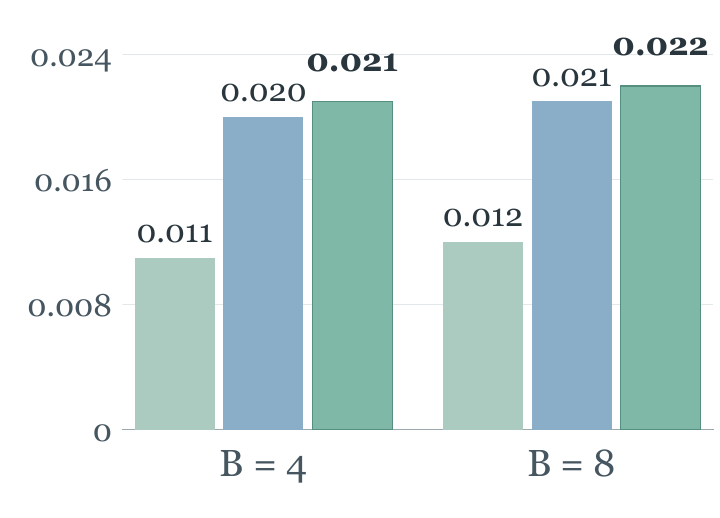}
\caption{Llama-3.1-8B}
\label{fig:matched_budget_llama}
\end{subfigure}
\caption{TLAR complements the strongest retrieval baseline under matched verification budgets on code debugging. In each budget group, the bars from left to right represent TLAR (light green), the strongest single-source retriever (blue), and their budgeted union (dark green). Bar heights show point estimates of additional accepted tokens per position over SmallDraft; labels give their values.}
\label{fig:matched_budget}
\end{figure}

\begin{figure}[t]
\centering
\begin{subfigure}[t]{0.32\linewidth}
\includegraphics[width=\linewidth]{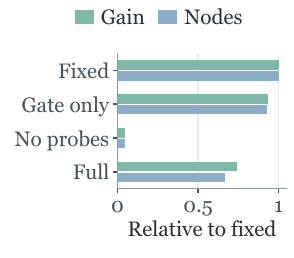}
\vspace{-2mm}
\caption{Qwen3-8B}
\end{subfigure}\hfill
\begin{subfigure}[t]{0.32\linewidth}
\includegraphics[width=\linewidth]{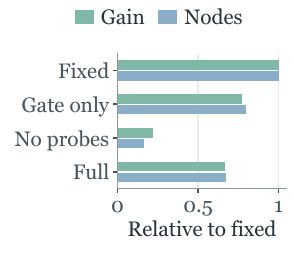}
\caption{Mistral-24B}
\end{subfigure}\hfill
\begin{subfigure}[t]{0.32\linewidth}
\includegraphics[width=\linewidth]{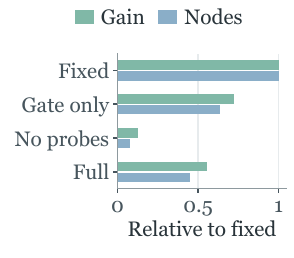}
\caption{Llama-3.1-8B}
\end{subfigure}
\caption{Controller choices change both accepted gain and verification work. Gain is the increment over SmallDraft; nodes are distinct retrieval prefixes. Both are normalized by the fixed-gate, fixed-width replay, not the branch-count costs in Appendix~\ref{app:main_cost}.}
\vspace{-2mm}
\label{fig:adaptive}
\end{figure}

\begin{figure}[t]
\centering
\begin{subfigure}[t]{0.32\linewidth}
\includegraphics[width=\linewidth]{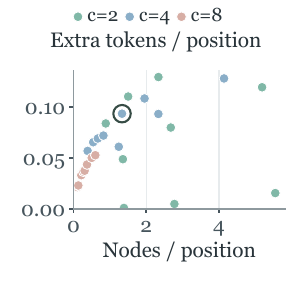}
\caption{Qwen3-8B}
\end{subfigure}\hfill
\begin{subfigure}[t]{0.32\linewidth}
\includegraphics[width=\linewidth]{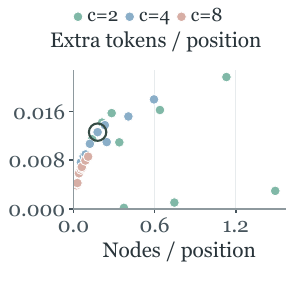}
\caption{Mistral-24B}
\end{subfigure}\hfill
\begin{subfigure}[t]{0.32\linewidth}
\includegraphics[width=\linewidth]{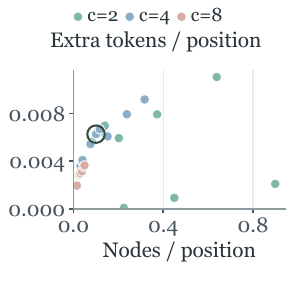}
\caption{Llama-3.1-8B}
\end{subfigure}
\caption{Candidate settings change recovered gain and verification work. All 27 combinations of \(c\in\{2,4,8\}\), \(\epsilon\in\{0,1,2\}\), and \(k\in\{2,4,8\}\) are shown. Color indicates context length; the circled point is \(c=4,\epsilon=1,k=4\).}
\vspace{-2mm}
\label{fig:revision_grid}
\end{figure}

\subsection{RQ3: How Does Adaptive Retrieval Change Gain and Cost?}
\label{sec:gain_cost}

We compare the full controller with a fixed gate and width, adaptive activation alone, and a controller without periodic probes, using the same replay and distinct-prefix node counter. We normalize both added acceptance over
SmallDraft and nodes by the fixed configuration in Figure~\ref{fig:adaptive}. Separately, we vary context length, mismatch tolerance, and candidate width
over 27 settings per model in Figure~\ref{fig:revision_grid}. This grid measures
recovery at SmallDraft first-token misses.
For the controller comparison, both quantities equal one for the fixed configuration. A lower node value means less retrieval-side verification work; a lower gain value means fewer additional accepted tokens. Reading both is
necessary: lower cost alone can also result from rarely retrieving. The grid instead holds the evaluation rule fixed and changes the candidates, so its highest-gain point is a setting of the same retrieval method.
The controller reduces node cost but also gives up accepted tokens. Gain per node improves for Qwen and Llama. Removing probes lowers gain for every model, supporting their role in finding useful content after retrieval becomes inactive. Candidate settings offer a separate trade-off: the default uses fewer nodes than the highest-gain setting in each
model's grid, but recovers fewer tokens. The grid changes available candidates; the controller changes when and how much retrieval is used. We provide exact values and sensitivity tests in Appendix~\ref{app:revision_results}.
\begin{wraptable}{r}{0.60\textwidth}
\vspace{-6pt}
\centering
\small
\setlength{\tabcolsep}{2pt}
\captionsetup{
  justification=justified,
  singlelinecheck=false,
  margin=0pt
}
\caption{Qwen/B200 throughput (tokens/s) across output lengths.
Values are medians of five warmed runs; gains are relative to
SmallDraft. At 512 tokens, retrieval is inactive and the difference
captures controller/index overhead.}
\label{tab:length_scaling}
\begin{tabular}{@{}rrrrrrr@{}}
\toprule
& \multicolumn{3}{c}{Batch 1}
& \multicolumn{3}{c}{Batch 32} \\
\cmidrule(lr){2-4}\cmidrule(lr){5-7}
Tokens & SmallDraft & Union & Gain
       & SmallDraft & Union & Gain \\
\midrule
512
& 151.80 & 150.89 & $-0.6\%$
& 2528.0 & 2510.3 & $-0.7\%$ \\
1,024
& 143.20 & 155.09 & $+8.3\%$
& 2405.0 & 2587.8 & $+7.6\%$ \\
2,048
& 132.60 & 146.52 & $+10.5\%$
& 2222.0 & 2422.0 & $+9.0\%$ \\
4,096
& 119.40 & 134.21 & $+12.4\%$
& 1986.0 & 2200.5 & $+10.8\%$ \\
\bottomrule
\end{tabular}
\vspace{-6pt}
\end{wraptable}
At matched verification-node costs, recent-hit feedback yields higher accepted-token increments than random or periodic activation across all three targets. We show further details in Appendix Table~\ref{tab:controller_signal_ablation}.

\noindent\textbf{End-to-end evaluation.}
We evaluate the complete adaptive union tree from
Section~\ref{sec:method} for Qwen3-8B/Qwen3-0.6B and
Mistral-24B/Ministral-8B on one NVIDIA B200.
End-to-end timing includes retrieval, tree construction,
prefix deduplication, budgeted verification, controller updates, KV-cache updates, and sampling correction.
Relative to the same draft-model baseline, throughput improves by 8.3\% and 7.6\% for Qwen and 4.1\% and 3.5\% for Mistral at batch sizes 1 and 32, respectively.
Table~\ref{tab:length_scaling} shows that Qwen's relative gains increase with output length, reaching 12.4\% and 10.8\% at 4,096 tokens. Longer generation provides more positions for active retrieval beyond the 512-token gate, helping amortize history-maintenance costs.
Appendix~\ref{app:wallclock} provides execution settings
and repeated-run distributions.

\noindent \textbf{From acceptance to execution efficiency.}
Throughput depends on both the tokens committed per decoding event and the elapsed time per event. Additional accepted tokens increase progress, while retrieval, tree construction, verification, and memory movement contribute to execution time. Appendix~\ref{app:systems_efficiency} formalizes this relationship, calculates net time savings from the measured runtimes, and examines the implementation's resource demands and opportunities for overlap. It also develops a controlled component-evaluation framework linking candidate quality, execution cost, and serving performance, motivating retrieval policies that adapt to both recent reuse and system load.
\vspace{-1mm}
\section{Related Work}
\label{related}

\textbf{Model-based drafting.}
Speculative decoding uses a faster drafter to propose tokens and a target model to verify them~\citep{leviathan2023fast,chen2023accelerating}.
EAGLE and Medusa improve proposal quality through trained drafting components \citep{li2024eagle,cai2024medusa}, while SpecInfer, EAGLE-2, Sequoia, and
OPT-Tree improve tree construction and parallel verification
\citep{miao2024specinfer,li2024eagle2,chen2024sequoia,wang2024opttree}.
These methods improve how model-generated candidates are produced or verified, but do not test whether the generated trajectory provides a distinct candidate source beyond the same draft model. TLAR addresses this question by measuring
the additional accepted tokens supplied by trajectory retrieval after SmallDraft has already proposed its candidates.

\textbf{Retrieval-based drafting.}
Retrieval methods obtain candidates from the prompt, external data, or generated
text~\citep{somasundaram2025pld+,he2024rest,fu2024lookahead,
hu2025sam,oliaro2026suffixdecoding}.
STAND exploits redundancy in reasoning trajectories
\citep{song2025stand}, and RASD combines retrieval and model-generated trees before verification~\citep{quan2025rasd}.
However, aggregate retrieval acceptance does not establish that useful tokens are specific to the current trajectory, and combining retrieval with a draft model does not show whether retrieval contributes candidates the drafter would
otherwise miss. TLAR isolates trajectory-specific reuse through cross-trajectory, shuffled-history, and unigram controls, then measures its marginal contribution beyond SmallDraft and retrieval baselines under matched
verification budgets.

\textbf{Adaptive speculative computation.}
Prior work adapts candidate length, lookahead, or draft-tree structure using predicted acceptance or draft-model confidence \citep{huang2024specdec++,mamou2024disco,
xiong2024dyspec,wang2024opttree}.
These policies optimize model-generated speculation, but do not use the realized marginal value of trajectory retrieval to decide whether retrieval should run or how many retrieved candidates should be verified. TLAR uses only
causally available outcomes from earlier target verification to control later retrieval activation and width. It connects source identification,
marginal utility beyond an existing drafter, and feedback-based allocation of verification work in one evaluation.

\vspace{-1mm}
\section{Conclusion}
\label{conclusion}

Generated trajectories serve as runtime memory for speculative decoding. Source comparisons reveal trajectory-specific reuse and temporal locality, motivating TLAR's retrieval from generated history and adaptation through recent verification feedback. TLAR adds accepted tokens beyond a draft model and complements strong retrieval baselines under shared verification budgets.
Its controller adjusts retrieval activation and candidate width to balance acceptance gains with verification cost.
The complete adaptive union improves end-to-end throughput,
with Qwen's relative gains increasing over the tested output lengths. These results connect trajectory reuse, candidate complementarity, and adaptive resource allocation to efficient inference.

\section*{Ethics Statement}

TLAR accelerates language-model inference through reuse of generated text. Exact target-model verification preserves the target distribution, including its potential to produce biased, harmful, or inaccurate content. Deployments should retain appropriate safety safeguards as generation becomes faster. Generated trajectories may contain sensitive information; retrieval buffers should follow the access controls, retention policies, and deletion requirements applied to the original interaction. Our trajectory-local design keeps retrieval within the current generation. Resource use also warrants attention: throughput improvements can reduce execution time for a fixed
workload, while energy consumption depends on hardware utilization and deployment scale. Responsible evaluation and release should document experimental conditions and protect sensitive data, credentials.

\section*{AI Use Statement}

AI-assisted tools were used solely for grammar checking and presentation-quality review, including the identification of sentence fragments, dangling words, and formatting inconsistencies. The authors reviewed and approved every resulting edit and retain full responsibility for the manuscript.

\section*{Reproducibility Statement}

The appendix documents model checkpoints, evaluation settings, retrieval configurations, verification budgets, controller parameters, and metric definitions. It also describes uncertainty calculations, causal-boundary and distribution-preservation audits, and the hardware, timing protocol, and repeated-run results for end-to-end evaluation. The accompanying anonymized code provides retrieval, candidate-tree construction, verification, adaptive control, and evaluation utilities. 

\setlength{\parskip}{5.8pt plus 2pt minus 0.3pt}
\setlength{\textfloatsep}{10pt plus 3pt minus 1pt}
\setlength{\floatsep}{8pt plus 2pt minus 1pt}
\setlength{\intextsep}{8pt plus 1pt minus 1pt}
\captionsetup{skip=5pt}
\captionsetup[subfigure]{skip=3pt}

\par
\endgroup

\bibliography{iclr2027_conference}
\bibliographystyle{iclr2027_conference}

\clearpage
\appendix
\makeatletter
\@ifundefined{FloatBarrier}{%
  \def\@fb@botlist{\@botlist}%
  \def\@fb@topbarrier{\suppressfloats[t]}%
  \def\FloatBarrier{\par\begingroup\let\@elt\relax
    \edef\@tempa{\@fb@botlist\@deferlist\@dbldeferlist}%
    \ifx\@tempa\@empty
    \else
      \ifx\@fltovf\relax
        \if@firstcolumn
          \clearpage
        \else
          \null\newpage\FloatBarrier
        \fi
      \else
        \newpage\let\@fltovf\relax\FloatBarrier
      \fi
    \fi
    \endgroup\@fb@topbarrier}%
}{}
\makeatother

\begingroup
\hypersetup{hidelinks}
\setlength{\parskip}{0pt}
\section*{Appendix Contents}
\pdfbookmark[0]{Appendix Contents}{appendix-contents}
\fontsize{9.5}{11.8}\selectfont
\newcommand{\appcontentssection}[2]{%
  \par\vspace{3pt}\noindent
  \hyperref[#1]{\textbf{\makebox[2.8em][l]{\ref*{#1}}#2}}%
  \nobreak\leaders\hbox{\kern1pt.\kern1pt}\hfill
  \nobreak\hyperref[#1]{\pageref*{#1}}\par}
\newcommand{\appcontentssubsection}[2]{%
  \noindent\hspace*{.9em}%
  \hyperref[#1]{\makebox[2.8em][l]{\ref*{#1}}#2}%
  \nobreak\leaders\hbox{\kern1pt.\kern1pt}\hfill
  \nobreak\hyperref[#1]{\pageref*{#1}}\par}
\appcontentssection{app:contents:01}{Related Work and Research Motivation}
\appcontentssubsection{app:contents:02}{Reasoning Length and Decoding Work}
\appcontentssubsection{app:contents:03}{Draft Construction and Exact Verification}
\appcontentssubsection{app:contents:04}{Text Sources as Draft Memory}
\appcontentssubsection{app:contents:05}{Feedback and the Allocation of Verification Work}
\appcontentssubsection{app:contents:06}{Trajectory Structure and the Role of TLAR}
\appcontentssection{app:contents:07}{Evaluation Protocol, Metrics, and Implementation}
\appcontentssubsection{app:contents:08}{Models, Tasks, and Experimental Snapshots}
\appcontentssubsection{app:contents:09}{Retrieval Sources and Shared Implementations}
\appcontentssubsection{app:contents:10}{Acceptance, Recovery, and Verification Cost}
\appcontentssubsection{app:contents:11}{Uncertainty and Result Provenance}
\appcontentssubsection{app:contents:12}{Implementation Flow}
\appcontentssection{app:contents:13}{Trajectory-Local Reuse and Source Controls}
\appcontentssubsection{app:contents:14}{Source Comparisons Across Models and Domains}
\appcontentssubsection{app:contents:15}{Repetition Audit and the Interpretation of Long Matches}
\appcontentssubsection{app:contents:16}{Same-Prompt, Different-Seed Pairing}
\appcontentssubsection{app:contents:17}{Temporal Dependence as a Control Signal}
\appcontentssection{app:contents:18}{Candidate Complementarity and Verification Budgets}
\appcontentssubsection{app:contents:19}{Acceptance Beyond SmallDraft}
\appcontentssubsection{app:contents:20}{Domain Variation and Draft-Miss Opportunities}
\appcontentssubsection{app:contents:21}{Complementarity Under Shared Node Caps}
\appcontentssubsection{app:contents:22}{Supplementary Candidate-Coverage Analysis}
\appcontentssection{app:contents:23}{Adaptive Retrieval and Verification Cost}
\appcontentssubsection{app:contents:24}{Candidate Settings and the Gain--Cost Frontier}
\appcontentssubsection{app:contents:25}{Activation, Width, and Periodic Probes}
\appcontentssubsection{app:contents:26}{Feedback Aligned with Incremental Acceptance}
\appcontentssubsection{app:contents:27}{Context Matching Under the Same Controller}
\appcontentssection{app:contents:28}{Causal Construction and Exact Verification}
\appcontentssubsection{app:contents:29}{Historical Candidate Construction}
\appcontentssubsection{app:contents:30}{Exact Tree Verification and Distribution Preservation}
\appcontentssubsection{app:contents:31}{Event-Driven Replay and Controller Updates}
\appcontentssection{app:contents:32}{End-to-End Execution and Length Scaling}
\appcontentssubsection{app:contents:33}{Execution Configuration and Target--Draft Pairs}
\appcontentssubsection{app:contents:34}{Throughput of the Complete Union Tree}
\appcontentssubsection{app:contents:35}{Repeated-Run Variation and Paired Summaries}
\appcontentssubsection{app:systems_efficiency}{From Candidate Acceptance to Execution Efficiency}
\appcontentssubsection{app:contents:37}{Archived Model-Based Drafting Reference}
\appcontentssection{app:contents:38}{Discussion: Trajectory Memory as an Inference Resource}
\appcontentssubsection{app:contents:39}{Generation Creates a Memory Whose Value Depends on the Drafter}
\appcontentssubsection{app:contents:40}{A Verification Budget Makes Candidate Composition Central}
\appcontentssubsection{app:contents:41}{Feedback Should Track the Value Added to the Existing System}
\appcontentssubsection{app:contents:42}{Exact Verification Supports Flexible Memory Search}
\appcontentssubsection{app:contents:43}{Longer Reasoning Changes Both the Memory and Its Economics}
\appcontentssubsection{app:contents:44}{An Evaluation Chain for Memory-Based Inference}
\endgroup
\clearpage

\par
\newsavebox{\appendixtablebox}
\newcommand{\fitappendixtable}[1]{%
  \sbox{\appendixtablebox}{#1}%
  \ifdim\wd\appendixtablebox>\linewidth
    \resizebox{\linewidth}{!}{\usebox{\appendixtablebox}}%
  \else\usebox{\appendixtablebox}\fi}
\makeatletter
\setlength{\@fptop}{0pt}
\setlength{\@fpsep}{14pt}
\setlength{\@fpbot}{0pt plus 1fil}
\makeatother
\renewcommand{\topfraction}{0.95}
\renewcommand{\bottomfraction}{0.90}
\renewcommand{\textfraction}{0.05}
\setcounter{topnumber}{5}
\setcounter{bottomnumber}{5}
\setcounter{totalnumber}{8}
\setlength{\textfloatsep}{8pt plus 2pt minus 2pt}
\setlength{\intextsep}{8pt plus 2pt minus 2pt}
\setlength{\floatsep}{8pt plus 2pt minus 2pt}

\setlength{\parskip}{3pt plus 1pt minus .5pt}
\renewcommand{\arraystretch}{1.08}
\renewcommand{\floatpagefraction}{0.88}

\AddToHook{cmd/section/before}{\FloatBarrier}
\raggedbottom

\section{Related Work and Research Motivation}
\label{app:contents:01}
\label{app:related_work}

Section~\ref{related} places TLAR among speculative decoding methods. We develop
that connection through the sequence that motivates the method: longer reasoning
creates a decoding workload, speculative verification accelerates token generation,
retrieval supplies candidate continuations, and feedback allocates verification work.

\subsection{Reasoning Length and Decoding Work}
\label{app:contents:02}
Chain-of-thought prompting elicits intermediate reasoning steps~\citep{wei2022chain}.
Self-consistency and repeated sampling allocate computation across reasoning paths
\citep{wang2022self,brown2024large}, while test-time scaling extends the amount of
reasoning used to produce an answer~\citep{muennighoff2025s1}. These approaches make
the execution cost of long trajectories a central inference problem.
Acceleration can act at several levels. SpecReason delegates selected reasoning
steps using semantic acceptance~\citep{pan2026specreason}. Speculative search and
speculative chain-of-thought methods organize parallel reasoning exploration
\citep{wang2025accelerating,wang2026efficient,shi-etal-2025-speccot}; multi-sample
speculative decoding shares work across sampled outputs~\citep{li2025speculative}.
TLAR acts at the token level, retrieving continuations and applying exact target
verification. This choice connects reasoning trajectories to lossless decoding.

\subsection{Draft Construction and Exact Verification}
\label{app:contents:03}
Blockwise decoding proposes several future tokens and retains a verified prefix
\citep{stern2018blockwise}. Speculative sampling combines a cheaper drafter with
parallel target verification and a sampling correction that preserves the target
distribution~\citep{leviathan2023fast,chen2023accelerating}. Surveys organize the
resulting design space around drafting and verification~\citep{xia2024unlocking}.

Medusa introduces prediction heads~\citep{cai2024medusa}; EAGLE and its successors
use feature prediction and dynamic candidate trees
\citep{li2024eagle,li2024eagle2,li2025eagle3}. SpecInfer verifies candidate trees
\citep{miao2024specinfer}, and Sequoia and OPT-Tree optimize tree structure for
acceptance and hardware cost~\citep{chen2024sequoia,wang2024opttree}.
Ouroboros extends drafts phrase by phrase~\citep{zhao2024ouroboros}; self-speculation,
early exit, and layer skipping reuse the target model's computation
\citep{zhang2024draft,elhoushi2024layerskip}. Lookahead Decoding builds candidates
through parallel Jacobi-style iteration~\citep{fu2024lookahead}.
TLAR supplies an additional candidate source to this draft--verify architecture:
continuations already present in the generated trajectory.

\subsection{Text Sources as Draft Memory}
\label{app:contents:04}
LLMA copies and verifies reference spans~\citep{yang2023inference}. REST retrieves
from an external datastore~\citep{he2024rest}, PLD+ exploits repeated prompt and
generation text~\citep{somasundaram2025pld+}, RAPID combines retrieval with
long-context speculation~\citep{chen2025rapid}, and NEST uses nearest-neighbor
text for generation and attribution~\citep{li2024nearest}.
Generated and cached sequences provide another memory source: SuffixDecoding
uses suffix-tree statistics~\citep{oliaro2026suffixdecoding}, SAM-Decoding uses
suffix automata~\citep{hu2025sam}, Token Recycling reuses candidate transitions
\citep{luo2025turning}, and STAND uses stochastic n-gram statistics from reasoning
trajectories~\citep{song2025stand}.
These systems motivate a source-level question: which continuations become useful
because they belong to the trajectory currently being generated? Section~\ref{sec:exploratory}
addresses this question with cross-trajectory, shuffled-history, unigram, and
same-prompt comparisons. Sections~\ref{sec:main_results} and~\ref{sec:rq2} then
measure the extra acceptance supplied by that source alongside existing drafters.

\subsection{Feedback and the Allocation of Verification Work}
\label{app:contents:05}
The benefit of a candidate depends on its acceptance and its verification cost.
SpecDec++ predicts acceptance~\citep{huang2024specdec++}; DISCO adapts lookahead
using draft-model signals~\citep{mamou2024disco}; EAGLE-2 and DySpec grow trees
using draft confidence or distributional information~\citep{li2024eagle2,xiong2024dyspec}.
OPT-Tree and Sequoia optimize expected acceptance and tree cost
\citep{wang2024opttree,chen2024sequoia}, while token-driven calibration adapts
speculation during execution~\citep{gautam2025token}.

TLAR uses completed target-verification events to control retrieval activation
and width. Periodic probes refresh the signal after retrieval becomes inactive.
The controller ablations in Appendix~\ref{app:w7} connect this feedback to
incremental acceptance, and Appendix~\ref{app:causal_validation} specifies how
the feedback is generated along the executed decoding path.

\subsection{Trajectory Structure and the Role of TLAR}
\label{app:contents:06}
ReasonOps identifies recurring reasoning operations~\citep{lee2026reasonops}, and
analyses of overthinking describe clustered backtracking~\citep{rezazadeh2026shape}.
These observations motivate studying structure within a trajectory. TLAR measures
that structure through context matches and exactly accepted continuations.
Its verifier preserves the target distribution; speculative contrastive decoding
and fuzzy speculation explore altered decoding objectives and accuracy--runtime
trade-offs~\citep{yuan2024speculative,holsman2025fuzzy}.

RASD provides a closely related hybrid architecture: it prunes retrieval candidates
using draft probabilities and merges retrieval and model-generated trees
\citep{quan2025rasd}. Building on retrieval and tree verification, TLAR connects
source identification, added acceptance, and feedback-driven resource allocation.
Table~\ref{tab:closest_positioning} summarizes how candidate sources and control
signals fit together. The remaining appendices follow that experimental chain.

\begin{table}[!htbp]
\centering\scriptsize\setlength{\tabcolsep}{2.7pt}
\caption{Candidate sources and control mechanisms in closely related systems. Feedback control uses earlier target-verification outcomes to change later retrieval work. Dashes indicate absent components.}
\label{tab:closest_positioning}
\fitappendixtable{\begin{tabular}{llllll}
\toprule
Method & Candidate source & Draft-tree fusion & Exact verification &
Feedback control & Selection principle \tabularnewline
\midrule
STAND & reasoning trajectories & --- & yes & --- &
stochastic n-gram sampling \tabularnewline
SuffixDecoding & cached/generated suffixes & --- & yes & tree structure &
suffix statistics \tabularnewline
RASD & retrieval database & yes & yes & --- &
probability pruning and tree fusion \tabularnewline
TLAR & current trajectory & yes & yes & yes &
source isolation and causal budget allocation \tabularnewline
\bottomrule
\end{tabular}}
\end{table}

\section{Evaluation Protocol, Metrics, and Implementation}
\label{app:contents:07}
\label{app:protocol_section}
\label{app:revision_results}

This section supplies the common setup for Section~\ref{sec:setup}. We first define
the tasks and candidate sources, then the acceptance and cost counters, and finally
the estimation and implementation procedures. These definitions connect each
result table to its reference drafter and evaluated positions.

\subsection{Models, Tasks, and Experimental Snapshots}
\label{app:contents:08}
\label{app:protocol}
The main suite contains 100 code-debugging, 24 mathematical-reasoning, and 24
open-ended-writing problems per target. Target generation uses temperature $0.6$,
top-$p=0.95$, and a maximum of 4,096 new tokens. Within each target, retrieval
comparisons share fixed trajectories and SmallDraft predictions.
Table~\ref{tab:checkpoints} gives the target--draft pairs.

\begin{table}[!htbp]
\centering\small
\caption{Target--draft pairs for the offline comparisons and the B200 execution analysis.}
\label{tab:checkpoints}
\begin{tabular}{ll}\toprule
Target & SmallDraft checkpoint \\\midrule
Qwen3-8B & \texttt{Qwen/Qwen3-0.6B} \\
Mistral-24B & \texttt{mistralai/Ministral-8B-Instruct-2410} \\
Llama-3.1-8B & \texttt{meta-llama/Llama-3.2-1B-Instruct} \\\bottomrule
\end{tabular}
\end{table}

Code prompts contain SWE-bench Verified issue descriptions and repository metadata
\citep{jimenez2024swebench}; gold patches and hidden tests remain in the evaluation
data. The MATH test problems span algebra, counting/probability, geometry, and
number theory~\citep{hendrycks2021math}. WritingPrompts supplies the open-ended
prompts, with reference stories retained as evaluation data~\citep{fan2018writing}.
Examples are sorted by canonical identifier and sampled with seed 20260217 before
retrieval evaluation. The same-prompt comparison uses two generation seeds for
each of 24 code-debugging problems.

Table~\ref{tab:dataset_versions} records the two experimental snapshots. The
historical source audit includes DeepSeek and five writing prompts; the main
suite includes Mistral and 24 writing prompts. Their roles are source analysis
and the expanded utility evaluation, respectively.

\begin{table}[!htbp]
\centering\small\setlength{\tabcolsep}{3pt}
\caption{Experimental snapshots and their roles. Code, Math, and Open give problem counts per target.}
\label{tab:dataset_versions}
\fitappendixtable{\begin{tabular}{llrrrl}
\toprule
Snapshot & Models & Code & Math & Open & Role \tabularnewline
\midrule
Legacy source audit & Qwen, DeepSeek, Llama &
100 & 24 & 5 & nine-cell source diagnostic in Appendix~\ref{app:selfcopy} \tabularnewline
Expanded main suite & Qwen, Mistral, Llama &
100 & 24 & 24 & RQ1--RQ3 and B200 evaluation \tabularnewline
\bottomrule
\end{tabular}}
\end{table}

\subsection{Retrieval Sources and Shared Implementations}
\label{app:contents:09}
PLD searches the current prompt. The REST proxy indexes the other prompts in the
trace pool, separated by sentinel tokens; a 100-trace pool supplies 99 external
prompt entries for each query trajectory. The SAM proxy combines exact history
retrieval with this prompt datastore. STAND-style and SuffixDecoding candidates
use the common ranking, tree, and verification interfaces summarized in
Table~\ref{tab:baseline_scope}. This shared implementation supports comparisons
of candidate sources under the same evaluation rules.

\begin{table}[!htbp]
\centering\scriptsize\setlength{\tabcolsep}{3pt}
\caption{Retrieval sources and components implemented in the common evaluation harness.}
\label{tab:baseline_scope}
\fitappendixtable{%
\begin{tabular}{lp{.19\linewidth}p{.31\linewidth}p{.31\linewidth}}
\toprule
Displayed label & Search source & Implemented components &
Evaluation role \tabularnewline
\midrule
PLD & current prompt & exact n-gram lookup, $k=4,d=4$ &
prompt-source candidate comparison \tabularnewline
REST proxy & other prompts & external prompt index, exact context, sentinel boundaries &
shared-harness prompt-datastore proxy \tabularnewline
SAM proxy & history and other prompts & exact suffix lookup and union in the shared tree &
shared-harness history/datastore proxy \tabularnewline
STAND-style & generated history & stochastic n-gram candidates, common gate and budget &
shared-harness STAND-style candidates \tabularnewline
SuffixDecoding & generated/cache suffixes & exact suffix candidates and prefix tree &
common verifier and budgets; offline accepted-token comparison \tabularnewline
\bottomrule
\end{tabular}}
\end{table}

\subsection{Acceptance, Recovery, and Verification Cost}
\label{app:contents:10}
\label{app:metrics}
Following Section~\ref{prelim}, let $M_t=\mathbf{1}[a_{\mathrm{small}}(t)=0]$,
let $g_t$ indicate an active history gate, and let $a_{\mathrm{retr}}(t)$ be the
longest exactly accepted retrieval continuation. Let $N_t$ count retrieval work
under the counter specified for an experiment. The first-token-miss analysis uses
\[
 A_{\mathrm{miss}}=\frac{\sum_t M_tg_t}{\sum_t M_t},\qquad
 R=\frac{\sum_t M_tg_t\mathbf{1}[a_{\mathrm{retr}}(t)\ge1]}{\sum_t M_tg_t},
\]
\[
 G_{\mathrm{miss}}=\frac{1}{T}\sum_t M_tg_ta_{\mathrm{retr}}(t),\qquad
 C_{\mathrm{miss}}=\frac{1}{T}\sum_t M_tg_tN_t.
\]
Here $A_{\mathrm{miss}}$ is the active fraction of draft failures, recorded as
\texttt{active\_rate\_over\_miss}. Recovery $R$ uses active failures as its
denominator; $G_{\mathrm{miss}}$ and $C_{\mathrm{miss}}$ average over all evaluated
positions. Their ratio measures recovered tokens per verification node.

The hybrid comparison also credits retrieval for extending an accepted SmallDraft
prefix. Its acceptance and cost are
\[
 \Delta=\frac{1}{T}\sum_t\left[\max\{a_{\mathrm{small}}(t),g_ta_{\mathrm{retr}}(t)\}
 -a_{\mathrm{small}}(t)\right],\qquad
 C_{\mathrm{hyb}}=\frac{1}{T}\sum_t g_tN_t.
\]
This definition connects candidate usefulness directly to the existing drafter.
The fixed-candidate hybrid and static grid count summed branch lengths, including
shared prefixes. The adaptive replay, baseline-addition analysis, and budgeted
union count distinct prefix nodes. Each gain--cost comparison uses its experiment's
counter. The budgeted comparison gives every method the same maximum of $B$
distinct retrieval nodes per position.

\subsection{Uncertainty and Result Provenance}
\label{app:contents:11}
\label{app:estimation}
Trajectory-cluster bootstrap resamples trajectory identifiers with replacement,
retains all positions from each sampled trajectory, and recomputes the pooled
metric. We use 2,000 resamples and report the 2.5th and 97.5th percentiles.
This procedure preserves dependence among positions within a trajectory.
Same-prompt source comparisons pair the two rollouts by problem identifier.

The tables report experimental measurements. Dagger-marked entries in
Table~\ref{tab:main-full} identify updated experimental summaries. The reported
acceptance intervals and node-cost summaries accompany the corresponding
configurations. Table~\ref{tab:matched_budget} reports point estimates.
Appendix~\ref{app:wallclock} gives the B200 measurements and repeated-run
summaries, including the calculation of paired throughput intervals.

\subsection{Implementation Flow}
\label{app:contents:12}
The scripts in Table~\ref{tab:codebase} follow the evaluation pipeline: construct
prompts, generate trajectories, record draft predictions, compare candidate sources,
and aggregate acceptance and cost. Context matching selects historical locations;
the \texttt{accepted\_len} routine checks continuation tokens exactly until the
first mismatch or depth cap. Appendix~\ref{app:causal_validation} connects this
replay computation to incremental history insertion and executed decoding events.

\begin{table}[!htbp]
\centering
\small
\caption{Evaluation scripts and their roles in the data-to-metric pipeline.}
\label{tab:codebase}
\fitappendixtable{%
\begin{tabular}{p{.40\linewidth}p{.55\linewidth}}
\toprule
\textbf{Script} & \textbf{Role} \\
\midrule
\texttt{build\_code\_debug\_pool.py} & Constructs the code-debugging prompt pool. \\
\texttt{generate\_traces.py} & Generates target-model trajectories under the decoding configuration above. \\
\texttt{analyze\_draft\_hits.py} & Teacher-forces the small draft model over each trajectory; records $\text{hit}_{\text{draft}}(t)$. \\
\texttt{analyze\_selfcopy.py} & Computes within/cross/shuffled/unigram $E_{\text{exact}}$ (Section~\ref{sec:exploratory}). \\
\texttt{analyze\_tree\_hybrid.py} & Computes TLAR and hybrid (TLAR + SmallDraft) accepted lengths per position; evaluates individual retrieval trees. \\
\texttt{compare\_retrieval\_baselines.py} & Shared retrieval-baseline harness; source of the static grid and union hybrid. \\
\texttt{bootstrap\_metrics.py} & Trajectory-cluster bootstrap over the outputs of the scripts above. \\
\texttt{summarize\_analysis.py} & Computes development-time aggregate summaries. \\
\bottomrule
\end{tabular}%
}
\end{table}

\section{Trajectory-Local Reuse and Source Controls}
\label{app:contents:13}
\label{app:selfcopy}

Section~\ref{sec:exploratory} asks where reusable continuations originate. We
develop the evidence from source comparisons to repetition auditing, same-prompt
pairing, and temporal dependence. Each step explains a design choice in TLAR:
the memory source, the interpretation of accepted spans, and the feedback signal.

\subsection{Source Comparisons Across Models and Domains}
\label{app:contents:14}
Table~\ref{tab:source_full} supplies the Qwen, Mistral, and Llama values plotted
in Figure~\ref{fig:selfcopy3x3}. Within searches the generated prefix; Cross uses
another trajectory with the same available history length; Shuffled reorders
history blocks; Unigram uses token frequencies. Within, Cross, and Shuffled
share a matching rule, making the source the experimental variable.

\begin{table}[!htbp]
\centering\small
\setlength{\tabcolsep}{4pt}
\caption{Source means underlying Figure~\ref{fig:selfcopy3x3}. Entries are exactly accepted tokens; Unigram is the strongest control in each row.}
\label{tab:source_full}
\begin{tabular}{llrrrr}
\toprule
Task & Model & Within & Cross & Shuffled & Unigram \\
\midrule
Code debugging & Qwen3-8B & 1.2184 & 0.0139 & 0.0161 & 0.0658 \\
Code debugging & Mistral-24B & 2.7266 & 0.0069 & 0.0326 & 0.0585 \\
Code debugging & Llama-3.1-8B & 1.0612 & 0.0046 & 0.0236 & 0.0539 \\
\midrule
Mathematics & Qwen3-8B & 1.0620 & 0.0037 & 0.0155 & 0.0627 \\
Mathematics & Mistral-24B & 1.1888 & 0.0039 & 0.0268 & 0.0713 \\
Mathematics & Llama-3.1-8B & 2.2378 & 0.0141 & 0.0539 & 0.0761 \\
\midrule
Open-ended & Qwen3-8B & 0.0274 & 0.0016 & 0.0003 & 0.0535 \\
Open-ended & Mistral-24B & 0.1849 & 0.0003 & 0.0011 & 0.0435 \\
Open-ended & Llama-3.1-8B & 0.3283 & 0.0020 & 0.0121 & 0.0561 \\
\bottomrule
\end{tabular}

\end{table}

Current history yields higher accepted lengths than the controls in code and
mathematics. Open-ended writing has smaller accepted lengths and a different
ordering for Qwen: Within is $0.0274$ and Unigram is $0.0535$. The historical
Qwen/DeepSeek/Llama snapshot in Table~\ref{tab:selfcopy-full} adds the archived
intervals and within-to-control ratios. Unigram is the strongest control in all
nine cells, and the ratios range from $0.5\times$ to $29.4\times$.

\begin{table}[!htbp]
\centering\small
\setlength{\tabcolsep}{3.5pt}
\caption{Historical source snapshot: mean exactly accepted length and archived 95\% bootstrap intervals. Lift is Within divided by Unigram. The measurement cap is 32 tokens. $\dagger$ marks the cleaned Llama mathematics cell after removal of seven flagged trajectories.}
\label{tab:selfcopy-full}
\fitappendixtable{%
\begin{tabular}{lccccc}
\toprule
\textbf{Model} & \textbf{Within} & \textbf{Cross} & \textbf{Shuffled} & \textbf{Unigram} & \textbf{Lift} \\
\midrule
\multicolumn{6}{l}{\textit{Code-debugging}} \\
Qwen3-8B & \shortstack{1.2184\\{\scriptsize [1.1543, 1.2844]}} & \shortstack{0.0139\\{\scriptsize [0.0113, 0.0165]}} & \shortstack{0.0161\\{\scriptsize [0.0139, 0.0187]}} & \shortstack{0.0658\\{\scriptsize [0.0631, 0.0686]}} & 18.5$\times$ \\
DeepSeek-R1-Qwen-7B & \shortstack{0.6599\\{\scriptsize [0.6038, 0.7146]}} & \shortstack{0.0171\\{\scriptsize [0.0147, 0.0197]}} & \shortstack{0.0109\\{\scriptsize [0.0092, 0.0128]}} & \shortstack{0.0756\\{\scriptsize [0.0733, 0.0780]}} & 8.7$\times$ \\
Llama-3.1-8B & \shortstack{1.0612\\{\scriptsize [0.8361, 1.3284]}} & \shortstack{0.0046\\{\scriptsize [0.0022, 0.0080]}} & \shortstack{0.0236\\{\scriptsize [0.0135, 0.0351]}} & \shortstack{0.0539\\{\scriptsize [0.0511, 0.0568]}} & 19.7$\times$ \\
\midrule
\multicolumn{6}{l}{\textit{Mathematical reasoning}} \\
Qwen3-8B & \shortstack{1.0620\\{\scriptsize [0.9294, 1.2131]}} & \shortstack{0.0037\\{\scriptsize [0.0028, 0.0047]}} & \shortstack{0.0155\\{\scriptsize [0.0119, 0.0196]}} & \shortstack{0.0627\\{\scriptsize [0.0574, 0.0687]}} & 16.9$\times$ \\
DeepSeek-R1-Qwen-7B & \shortstack{0.8183\\{\scriptsize [0.6486, 1.0127]}} & \shortstack{0.0025\\{\scriptsize [0.0018, 0.0032]}} & \shortstack{0.0142\\{\scriptsize [0.0106, 0.0187]}} & \shortstack{0.0742\\{\scriptsize [0.0670, 0.0837]}} & 11.0$\times$ \\
Llama-3.1-8B$^\dagger$ & \shortstack{2.2378\\{\scriptsize [1.5816, 2.7283]}} & \shortstack{0.0141\\{\scriptsize [0.0062, 0.0222]}} & \shortstack{0.0539\\{\scriptsize [0.0187, 0.1165]}} & \shortstack{0.0761\\{\scriptsize [0.0616, 0.0898]}} & 29.4$\times$ \\
\midrule
\multicolumn{6}{l}{\textit{Open-ended writing}} \\
Qwen3-8B & \shortstack{0.0274\\{\scriptsize [0.0189, 0.0405]}} & \shortstack{0.0016\\{\scriptsize [0.0005, 0.0027]}} & \shortstack{0.0003\\{\scriptsize [0.0001, 0.0006]}} & \shortstack{0.0535\\{\scriptsize [0.0471, 0.0592]}} & 0.5$\times$ \\
DeepSeek-R1-Qwen-7B & \shortstack{0.1170\\{\scriptsize [0.0519, 0.1713]}} & \shortstack{0.0084\\{\scriptsize [0.0021, 0.0169]}} & \shortstack{0.0007\\{\scriptsize [0.0000, 0.0015]}} & \shortstack{0.0521\\{\scriptsize [0.0464, 0.0596]}} & 2.2$\times$ \\
Llama-3.1-8B & \shortstack{0.3283\\{\scriptsize [0.0675, 0.7089]}} & \shortstack{0.0020\\{\scriptsize [0.0000, 0.0042]}} & \shortstack{0.0121\\{\scriptsize [0.0025, 0.0204]}} & \shortstack{0.0561\\{\scriptsize [0.0480, 0.0682]}} & 5.9$\times$ \\
\bottomrule
\end{tabular}}
\end{table}

\subsection{Repetition Audit and the Interpretation of Long Matches}
\label{app:contents:15}
\label{app:audit}
Long accepted spans can arise from repeated output loops. To identify this source
of acceptance, the historical audit ranks trajectories by $E_{\mathrm{exact}}$
and inspects the ten highest-ranked trajectories per cell, or all five writing
trajectories. A trajectory is flagged when its mean is at least five tokens and
its maximum match reaches the 32-token measurement cap. A cell is classified as
repetition-dominated when over half of the inspected trajectories meet this rule.

\begin{table}[!htbp]
\centering\small
\setlength{\tabcolsep}{3.5pt}
\caption{Historical repetition audit. Counts describe the inspected high-repetition trajectories in each model--domain cell.}
\label{tab:audit-coverage}
\fitappendixtable{%
\begin{tabular}{llrr}
\toprule
\textbf{Domain} & \textbf{Model} & \textbf{Inspected} & \textbf{Flagged} \\
\midrule
code-debug100 & DeepSeek-R1-Qwen-7B & 10 & 0 \\
code-debug100 & Llama-3.1-8B & 10 & 0 \\
code-debug100 & Qwen3-8B & 10 & 0 \\
math & DeepSeek-R1-Qwen-7B & 10 & 0 \\
math & Llama-3.1-8B & 10 & 7 \\
math & Qwen3-8B & 10 & 0 \\
open-ended & DeepSeek-R1-Qwen-7B & 5 & 0 \\
open-ended & Llama-3.1-8B & 5 & 0 \\
open-ended & Qwen3-8B & 5 & 0 \\
\bottomrule
\end{tabular}}
\end{table}

The audit flags seven of the ten inspected Llama mathematics trajectories and
zero in each other inspected cell. Table~\ref{tab:audit-llama} gives the task-level
values. For example, task 22 repeats the same four-sentence geometric argument
more than a dozen times, ending with
\enquote{$\angle ABC=180^{\circ}-y-90^{\circ}=90^{\circ}-y$}.
This repeated passage produces long exact matches and illustrates why repetition
structure is relevant to interpreting retrieval acceptance.

\begin{table}[!htbp]
\centering
\small
\caption{The ten highest-repetition Llama mathematics trajectories. Flagged trajectories are removed when computing the cleaned source mean.}
\label{tab:audit-llama}
\begin{tabular}{cccc}
\toprule
\textbf{Rank} & \textbf{Task ID} & \textbf{Within $E_{\text{exact}}$} & \textbf{Flagged} \\
\midrule
1 & 3 & 16.28 & yes \\
2 & 22 & 16.24 & yes \\
3 & 7 & 11.67 & yes \\
4 & 16 & 8.87 & yes \\
5 & 18 & 6.77 & yes \\
6 & 4 & 5.89 & yes \\
7 & 6 & 5.61 & yes \\
8 & 8 & 3.91 & no \\
9 & 10 & 3.35 & no \\
10 & 2 & 2.61 & no \\
\bottomrule
\end{tabular}
\end{table}

Removing the seven flagged trajectories changes the Llama mathematics source
mean from $7.03$ to $2.2378$. Figure~\ref{fig:selfcopy3x3} and
Tables~\ref{tab:source_full} and~\ref{tab:selfcopy-full} use the cleaned mean.
The 32-token source-measurement cap quantifies reuse; TLAR's depth-four candidate
tree evaluates how that reuse supplies a practical draft.

\subsection{Same-Prompt, Different-Seed Pairing}
\label{app:contents:16}
\label{app:sameprompt}
The next comparison holds the problem fixed and changes the sampled trajectory. Each target has 24 paired problems. Qwen, Mistral and Llama use exact context matching and an alternative-history prefix of the same available length. Each row in Table~\ref{tab:sameprompt-full} compares own and other-seed
history under that target's protocol, as plotted in Figure~\ref{fig:g3c}.

\begin{table}[!htbp]
\centering\small
\caption{Same-prompt source comparisons on 24 paired code-debugging problems per model. Brackets give archived 95\% intervals; the alternative-source column specifies the matching protocol.}
\label{tab:sameprompt-full}
\begin{tabular}{llcc}\toprule
Model & Alternative source & Within & Other seed \\\midrule
Qwen3-8B & Prefix; exact context & 2.0814 [1.4340, 2.9241] & 1.2653 [0.9282, 1.6487] \\
Mistral-24B & Prefix; exact context & 1.4354 [1.1416, 1.7132] & 0.8614 [0.6644, 1.1190] \\
Llama-3.1-8B & Prefix; exact context & 0.8243 [0.3755, 1.4092] & 1.1134 [0.6498, 1.6673] \\\bottomrule
\end{tabular}
\end{table}

Qwen and Mistral have higher own-history means, while Llama has a higher other-seed
mean. Mistral's paired difference is $0.5740$ with interval $[0.1060,1.0065]$;
the table gives the marginal intervals for each source. These within-model
contrasts identify how trajectory identity affects candidate availability and
motivate the downstream complementarity tests in Appendix~\ref{app:results}.

\subsection{Temporal Dependence as a Control Signal}
\label{app:contents:17}
\label{app:temporal_reuse}
On code debugging, next-position hit probabilities after a hit are $0.76$, $0.86$,
and $0.81$ for Qwen, Mistral, and Llama. The overall probabilities on the same
position pairs are $0.46$, $0.50$, and $0.38$. A recent successful retrieval
provides information about the next position. Equation~\ref{eq:reuse_meter}
compresses this information into a running score, and the probes in
Section~\ref{method:controller} refresh that score as generation moves between
regions. Appendix~\ref{app:w7} measures the resulting gain--cost trade-off.

\section{Candidate Complementarity and Verification Budgets}
\label{app:contents:18}
\label{app:results}

The source tests establish candidate availability. The next question is how those
candidates extend an existing drafter. We follow RQ1 with the SmallDraft hybrid
comparison, examine domain and draft-miss effects, and then follow RQ2 with a
shared-budget comparison against a strong retrieval source.

\subsection{Acceptance Beyond SmallDraft}
\label{app:contents:19}
\label{app:main_cost}
Table~\ref{tab:main_hybrid_cost} expands Table~\ref{tab:main_hybrid} with verification
cost. The fixed configuration uses a 512-token history gate and $k=d=4$.
The union has the largest increment for each target, and its summed branch cost
also increases. Table~\ref{tab:main-full} gives the corresponding intervals and
identifies updated experimental summaries using the convention in
Appendix~\ref{app:estimation}.

\begin{table}[!htbp]
\centering\small
\setlength{\tabcolsep}{4pt}
\caption{Fixed-candidate hybrid acceptance and cost on 100 code-debugging tasks per target, with a 512-token gate and $k=d=4$. $e$ is SmallDraft acceptance and $\Delta$ its hybrid increment. Nodes sum branch lengths, including shared prefixes; the union sums both branch costs. Table~\ref{tab:main-full} gives intervals and revision markers.}
\label{tab:main_hybrid_cost}
\begin{tabular}{lrrrrrr}
\toprule
& \multicolumn{2}{c}{Qwen3-8B} & \multicolumn{2}{c}{Mistral-24B} & \multicolumn{2}{c}{Llama-3.1-8B} \\
\cmidrule(lr){2-3}\cmidrule(lr){4-5}\cmidrule(lr){6-7}
Added candidates & $\Delta$ & nodes & $\Delta$ & nodes & $\Delta$ & nodes \\
\midrule
SmallDraft alone & \multicolumn{2}{c}{$e=2.383$} & \multicolumn{2}{c}{$e=3.148$} & \multicolumn{2}{c}{$e=2.823$} \\
SuffixDecoding & 0.128 & 3.38 & 0.026 & 3.27 & 0.013 & 1.34 \\
STAND & 0.164 & 4.76 & 0.017 & 2.11 & 0.011 & 0.86 \\
\rowcolor{oursbg}
TLAR (fixed $k$) & 0.171 & 7.21 & 0.025 & 4.10 & 0.018 & 1.42 \\
STAND$\cup$TLAR & 0.228 & 11.98 & 0.032 & 6.22 & 0.026 & 2.28 \\
\bottomrule
\end{tabular}

\end{table}
\begin{table}[!htbp]
\centering\small
\caption{Full fixed-candidate hybrid measurements. Brackets give 95\% bootstrap intervals. Dagger-marked cells identify updated experimental summaries, as described in Appendix~\ref{app:estimation}.}
\label{tab:main-full}
\begin{tabular}{llcc}
\toprule
Model & Added retriever & $\Delta$ & Nodes / position \\
\midrule
Qwen3-8B & SuffixDecoding & \shortstack{0.1278\\{\scriptsize [0.1182, 0.1381]}} & \shortstack{3.3840\\{\scriptsize [3.1768, 3.5937]}} \\
Qwen3-8B & STAND & \shortstack{$0.1640^{\dagger}$\\{\scriptsize [0.1529, 0.1755]}} & \shortstack{$4.7635^{\dagger}$\\{\scriptsize [4.6361, 4.8879]}} \\
Qwen3-8B & TLAR & \shortstack{0.1709\\{\scriptsize [0.1606, 0.1809]}} & \shortstack{7.2146\\{\scriptsize [6.9291, 7.4923]}} \\
Qwen3-8B & STAND$\cup$TLAR & \shortstack{0.2280\\{\scriptsize [0.2141, 0.2422]}} & \shortstack{11.9781\\{\scriptsize [11.5626, 12.3594]}} \\
Mistral-24B & SuffixDecoding & \shortstack{0.0262\\{\scriptsize [0.0214, 0.0312]}} & \shortstack{3.2738\\{\scriptsize [2.5212, 4.1417]}} \\
Mistral-24B & STAND & \shortstack{$0.0170^{\dagger}$\\{\scriptsize [0.0109, 0.0233]}} & \shortstack{$2.1141^{\dagger}$\\{\scriptsize [1.7659, 2.4891]}} \\
Mistral-24B & TLAR & \shortstack{0.0253\\{\scriptsize [0.0189, 0.0324]}} & \shortstack{4.1033\\{\scriptsize [3.1826, 5.0426]}} \\
Mistral-24B & STAND$\cup$TLAR & \shortstack{0.0325\\{\scriptsize [0.0249, 0.0403]}} & \shortstack{6.2174\\{\scriptsize [4.9751, 7.4900]}} \\
Llama-3.1-8B & SuffixDecoding & \shortstack{$0.0130^{\dagger}$\\{\scriptsize [0.0090, 0.0173]}} & \shortstack{$1.3422^{\dagger}$\\{\scriptsize [1.1674, 1.5177]}} \\
Llama-3.1-8B & STAND & \shortstack{$0.0110^{\dagger}$\\{\scriptsize [0.0078, 0.0143]}} & \shortstack{$0.8609^{\dagger}$\\{\scriptsize [0.7485, 0.9717]}} \\
Llama-3.1-8B & TLAR & \shortstack{$0.0180^{\dagger}$\\{\scriptsize [0.0151, 0.0211]}} & \shortstack{$1.4209^{\dagger}$\\{\scriptsize [1.2036, 1.6435]}} \\
Llama-3.1-8B & STAND$\cup$TLAR & \shortstack{$0.0260^{\dagger}$\\{\scriptsize [0.0225, 0.0298]}} & \shortstack{$2.2817^{\dagger}$\\{\scriptsize [1.9761, 2.6201]}} \\
\bottomrule
\end{tabular}

\end{table}

The fixed-candidate TLAR increments are $0.1709$, $0.0253$, and $0.0180$ for Qwen,
Mistral, and Llama. Their meaning is incremental: they credit tokens that extend
the same SmallDraft's accepted prefix. The union's additional acceptance motivates
testing how both retrieval sources compete for a common node budget below.

\subsection{Domain Variation and Draft-Miss Opportunities}
\label{app:contents:20}
\label{app:boundary}
Table~\ref{tab:boundary} extends RQ1 to mathematics and writing. Recovery measures
success at active first-token misses; $G_{\mathrm{miss}}$ averages that recovery
over all positions; $\Delta$ also credits extensions of accepted draft prefixes.
For Mistral, source acceptance is higher in code ($2.7266$) than mathematics
($1.1888$), while the hybrid increment is higher in mathematics ($0.0610$) than
code ($0.0253$). This reversal connects source reuse to the coverage already
supplied by SmallDraft, as illustrated in Figure~\ref{fig:g3a}.

\begin{table}[!htbp]
\centering\small\setlength{\tabcolsep}{3pt}
\caption{Mathematics and writing results on 24 problems per domain and target. Brackets give archived 95\% bootstrap intervals. Metric definitions appear in Appendix~\ref{app:metrics}.}
\label{tab:boundary}
\fitappendixtable{\begin{tabular}{llccc}
\toprule
Domain & Model & Recovery & $G_{\mathrm{miss}}$ & $\Delta$ \\
\midrule
math & Qwen3-8B & \shortstack{0.1645\\{\scriptsize [0.1536, 0.1771]}} & \shortstack{0.0521\\{\scriptsize [0.0454, 0.0600]}} & \shortstack{0.0942\\{\scriptsize [0.0817, 0.1095]}} \\
math & Mistral-24B & \shortstack{0.1213\\{\scriptsize [0.1107, 0.1335]}} & \shortstack{0.0346\\{\scriptsize [0.0292, 0.0392]}} & \shortstack{0.0610\\{\scriptsize [0.0516, 0.0691]}} \\
math & Llama-3.1-8B & \shortstack{0.2718\\{\scriptsize [0.2336, 0.3056]}} & \shortstack{0.0271\\{\scriptsize [0.0220, 0.0317]}} & \shortstack{0.0557\\{\scriptsize [0.0445, 0.0657]}} \\
open-ended & Qwen3-8B & \shortstack{0.0141\\{\scriptsize [0.0094, 0.0209]}} & \shortstack{0.0060\\{\scriptsize [0.0047, 0.0074]}} & \shortstack{0.0082\\{\scriptsize [0.0057, 0.0112]}} \\
open-ended & Mistral-24B & \shortstack{0.0380\\{\scriptsize [0.0282, 0.0490]}} & \shortstack{0.0060\\{\scriptsize [0.0040, 0.0082]}} & \shortstack{0.0104\\{\scriptsize [0.0066, 0.0145]}} \\
open-ended & Llama-3.1-8B & \shortstack{0.0235\\{\scriptsize [0.0105, 0.0439]}} & \shortstack{0.0032\\{\scriptsize [0.0013, 0.0048]}} & \shortstack{0.0057\\{\scriptsize [0.0021, 0.0088]}} \\
\bottomrule
\end{tabular}
}
\end{table}

Draft failures reveal one route through which retrieval adds value.
Table~\ref{tab:crossmodel} reports their frequency and recovery before and after
history gating. Mistral has 11,401 first-token misses over 110,159 positions;
$0.4514$ of those misses pass the gate. Its recovery rises from $0.0946$ over
all misses to $0.1069$ over gated misses. The gate changes the evaluated set by
selecting positions with enough accumulated history.

\begin{table}[!htbp]
\centering\small
\caption{SmallDraft first-token misses and retrieval recovery on code debugging. Brackets give 95\% trajectory-cluster bootstrap intervals.}
\label{tab:crossmodel}
\begin{tabular}{lccc}\toprule
Model & Draft miss fraction & Ungated recovery & Gated recovery \\\midrule
Qwen3-8B & 0.207 [0.203, 0.212] & 0.193 [0.184, 0.202] & 0.222 [0.212, 0.232] \\
Mistral-24B & 0.1035 [0.0991, 0.1082] & 0.0946 [0.0789, 0.1121] & 0.1069 [0.0898, 0.1313] \\
Llama-3.1-8B & 0.145 [0.139, 0.152] & 0.051 [0.045, 0.058] & 0.089 [0.073, 0.106] \\\bottomrule
\end{tabular}
\end{table}

\paragraph{Expanded cross-domain evaluation.}
We extend the mathematics and open-ended writing evaluation to 100 problems per domain and target.
This expanded evaluation supports the cross-domain findings
in RQ1; the full recovery and acceptance breakdown appears here to connect those findings to draft-miss opportunities.
Table~\ref{tab:domain_100} preserves the domain ordering observed in the original evaluation: all three targets obtain larger acceptance increments in mathematics than in open-ended writing.
Within mathematics, Llama achieves the highest recovery rate, while Qwen obtains the largest increment over SmallDraft. Recovery measures retrieval success at active first-token misses.
Its contribution across the trajectory also depends on the
frequency of these opportunities and the accepted continuation length, while $\Delta$ additionally credits extensions of accepted draft prefixes. The expanded results reinforce the role of the existing drafter's coverage in determining the value of trajectory-local retrieval.

\begin{table}[!htbp]
\centering
\small
\setlength{\tabcolsep}{3pt}
\caption{Expanded cross-domain evaluation on 100 problems
per domain and target. Cells report point estimates with
intervals in brackets. Recovery measures success at active
draft misses; $G_{\mathrm{miss}}$ measures recovered tokens
per position; $\Delta$ measures additional acceptance over
SmallDraft. Metric definitions appear in
Appendix~\ref{app:metrics}.}
\label{tab:domain_100}
\begin{tabular*}{\linewidth}{
@{}ll@{\extracolsep{\fill}}ccc@{}
}
\toprule
Domain & Model & Recovery & $G_{\mathrm{miss}}$ & $\Delta$ \\
\midrule
Math & Qwen3-8B
& \shortstack{0.1682\\{\scriptsize [0.1628, 0.1741]}}
& \shortstack{0.0534\\{\scriptsize [0.0500, 0.0572]}}
& \shortstack{0.0971\\{\scriptsize [0.0907, 0.1038]}} \\
\addlinespace[3pt]
& Mistral-24B
& \shortstack{0.1187\\{\scriptsize [0.1134, 0.1248]}}
& \shortstack{0.0338\\{\scriptsize [0.0313, 0.0364]}}
& \shortstack{0.0594\\{\scriptsize [0.0549, 0.0636]}} \\
\addlinespace[3pt]
& Llama-3.1-8B
& \shortstack{0.2651\\{\scriptsize [0.2476, 0.2823]}}
& \shortstack{0.0264\\{\scriptsize [0.0239, 0.0288]}}
& \shortstack{0.0542\\{\scriptsize [0.0489, 0.0595]}} \\
\midrule
Open-ended & Qwen3-8B
& \shortstack{0.0153\\{\scriptsize [0.0127, 0.0184]}}
& \shortstack{0.0063\\{\scriptsize [0.0056, 0.0070]}}
& \shortstack{0.0087\\{\scriptsize [0.0073, 0.0102]}} \\
\addlinespace[3pt]
& Mistral-24B
& \shortstack{0.0356\\{\scriptsize [0.0309, 0.0408]}}
& \shortstack{0.0057\\{\scriptsize [0.0048, 0.0068]}}
& \shortstack{0.0098\\{\scriptsize [0.0079, 0.0118]}} \\
\addlinespace[3pt]
& Llama-3.1-8B
& \shortstack{0.0251\\{\scriptsize [0.0184, 0.0329]}}
& \shortstack{0.0034\\{\scriptsize [0.0025, 0.0043]}}
& \shortstack{0.0061\\{\scriptsize [0.0043, 0.0079]}} \\
\bottomrule
\end{tabular*}
\end{table}

\subsection{Complementarity Under Shared Node Caps}
\label{app:contents:21}
\label{app:matched_budget}
Table~\ref{tab:matched_budget} supplies the point estimates for
Figure~\ref{fig:matched_budget}. The strongest single source is STAND for Qwen
and Mistral and SuffixDecoding for Llama. Every method receives a cap of
$B\in\{4,8\}$ distinct retrieval nodes per position. The union merges the
two candidate sources, deduplicates prefixes, and prunes to that cap. Ranking
uses source-local recency and completed earlier verification events.

\begin{table}[!htbp]
\centering
\small
\setlength{\tabcolsep}{4pt}
\caption{Point estimates under shared verification-node caps, plotted in Figure~\ref{fig:matched_budget}. Best single is STAND for Qwen and Mistral and SuffixDecoding for Llama. The union combines that source with TLAR.}
\label{tab:matched_budget}
\begin{tabular}{lcccccc}
\toprule
& \multicolumn{3}{c}{$B=4$ nodes} &
\multicolumn{3}{c}{$B=8$ nodes} \\
Model & TLAR & Best single & Budgeted union &
TLAR & Best single & Budgeted union \\
\midrule
Qwen3-8B
& 0.128 & 0.168 & \textbf{0.191}
& 0.161 & 0.184 & \textbf{0.215} \\
Mistral-24B
& 0.022 & 0.026 & \textbf{0.029}
& 0.025 & 0.027 & \textbf{0.031} \\
Llama-3.1-8B
& 0.011 & 0.020 & \textbf{0.021}
& 0.012 & 0.021 & \textbf{0.022} \\
\bottomrule
\end{tabular}
\end{table}

The union has the largest increment in all six comparisons. Relative to the best
single source, its gains for Qwen, Mistral, and Llama are $13.7\%$, $11.5\%$, and
$5.0\%$ at $B=4$, and $16.8\%$, $14.8\%$, and $4.8\%$ at $B=8$.
These results connect the source evidence to better candidate composition under
a common verification allowance.

\subsection{Supplementary Candidate-Coverage Analysis}
\label{app:contents:22}
Table~\ref{tab:revision_plugin} identifies the additional continuations supplied
by TLAR to individual retrieval baselines. It retains each baseline's prefixes,
then appends previously absent TLAR prefixes up to the larger of the configured
width and the baseline prefix count. Acceptance and distinct-node increments are
reported together. This construction maps source overlap, while the preceding
shared-budget test evaluates competition for the available verification nodes.

\begin{table}[!htbp]
\centering\small
\caption{Supplementary coverage analysis on 100 code-debugging problems per target. Added tokens use the named retrieval baseline as reference; added nodes count the extra distinct prefixes. Baseline candidates are retained before new TLAR prefixes are appended.}
\label{tab:revision_plugin}
\begin{tabular}{llrrr}
\toprule
Model & Baseline & Added tokens & 95\% interval & Added nodes \\
\midrule
Qwen3-8B & PLD & 0.6341 & [0.6082, 0.6614] & 2.3890 \\
Qwen3-8B & REST & 0.6365 & [0.6097, 0.6623] & 2.3906 \\
Qwen3-8B & SAM-proxy & 0.1850 & [0.1798, 0.1900] & 0.9884 \\
Qwen3-8B & STAND & 0.0238 & [0.0225, 0.0250] & 0.2567 \\
Qwen3-8B & Suffix & 0.1879 & [0.1829, 0.1928] & 1.0006 \\
\addlinespace[2pt]
Mistral-24B & PLD & 0.5667 & [0.4295, 0.7136] & 1.3207 \\
Mistral-24B & REST & 0.5743 & [0.4406, 0.7265] & 1.3314 \\
Mistral-24B & SAM-proxy & 0.1002 & [0.0862, 0.1156] & 0.3672 \\
Mistral-24B & STAND & 0.0200 & [0.0161, 0.0242] & 0.1351 \\
Mistral-24B & Suffix & 0.1029 & [0.0879, 0.1195] & 0.3828 \\
\addlinespace[2pt]
Llama-3.1-8B & PLD & 0.2023 & [0.1696, 0.2368] & 0.5564 \\
Llama-3.1-8B & REST & 0.2022 & [0.1674, 0.2373] & 0.5552 \\
Llama-3.1-8B & SAM-proxy & 0.0571 & [0.0478, 0.0665] & 0.2213 \\
Llama-3.1-8B & STAND & 0.0131 & [0.0100, 0.0165] & 0.0894 \\
Llama-3.1-8B & Suffix & 0.0586 & [0.0492, 0.0681] & 0.2299 \\
\addlinespace[2pt]
\bottomrule
\end{tabular}

\end{table}

PLD and REST have larger increments than the history-based baselines. This pattern
relates the value of the added source to the text already searched by the baseline.
The next section uses feedback to allocate retrieval work across decoding positions.

\section{Adaptive Retrieval and Verification Cost}
\label{app:contents:23}
\label{app:w7}

RQ3 connects candidate quality to the amount of work spent obtaining accepted
tokens. We first vary the candidate set, then the activation policy, and finally
the feedback reward. This sequence links the retrieval mechanism in
Section~\ref{method:retrieval} to the controller in Section~\ref{method:controller}.

\subsection{Candidate Settings and the Gain--Cost Frontier}
\label{app:contents:24}
\begin{figure}[!htbp]\centering
\begin{subfigure}[t]{.32\linewidth}\includegraphics[width=\linewidth]{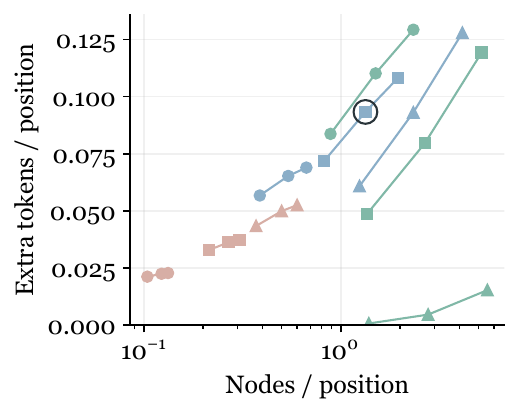}\caption{Qwen3-8B}\end{subfigure}\hfill
\begin{subfigure}[t]{.32\linewidth}\includegraphics[width=\linewidth]{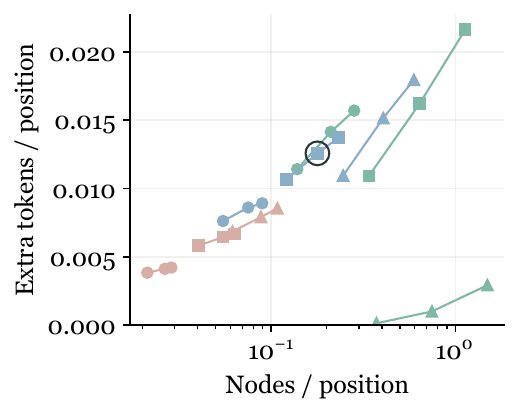}\caption{Mistral-24B}\end{subfigure}\hfill
\begin{subfigure}[t]{.32\linewidth}\includegraphics[width=\linewidth]{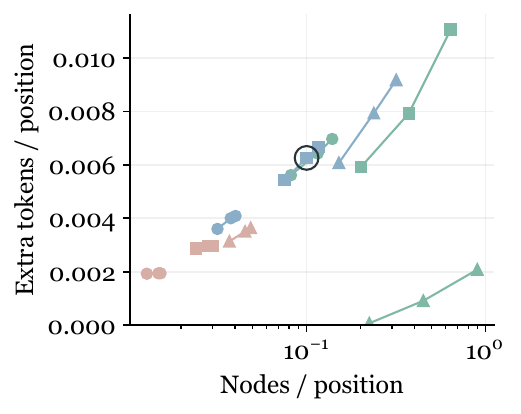}\caption{Llama-3.1-8B}\end{subfigure}\hfill
\caption{The 27 candidate settings per model. Green, blue, and pink indicate $c=2,4,8$; circles, squares, and triangles indicate $\epsilon=0,1,2$. Lines connect $k=2,4,8$; the outlined point is $(4,1,4)$. Horizontal axes are logarithmic, and panels use separate scales.}
\label{fig:w7-cost}\label{fig:w7-grid}
\end{figure}

The static grid contains all 27 combinations of $c\in\{2,4,8\}$,
$\epsilon\in\{0,1,2\}$, and $k\in\{2,4,8\}$ per target on 100 code-debugging
problems. Depth four and the 512-token gate remain fixed. The 81 archived settings
provide the full grid and the within-grid selections in
Tables~\ref{tab:w7-full} and~\ref{tab:w7-grid}. Figure~\ref{fig:w7-grid} shows
how context length, tolerance, and width move the gain--cost operating point.

\begin{table}[!htbp]
\centering\small\setlength{\tabcolsep}{3pt}
\caption{Selections from the archived static grid. $G=G_{\mathrm{miss}}$ measures recovered tokens, $C=C_{\mathrm{miss}}$ measures verification cost, and $G/C$ measures efficiency.}
\label{tab:w7-grid}
\fitappendixtable{\begin{tabular}{llcrrr}
\toprule
Model & Selection & $(c,\epsilon,k)$ & $G$ & $C$ & $G/C$ \\
\midrule
Qwen3-8B & Default & 4,1,4 & 0.0933 & 1.3344 & 0.0699 \\
Qwen3-8B & Largest observed gain & 2,0,8 & 0.1293 & 2.3356 & 0.0554 \\
Qwen3-8B & Highest gain per node & 8,0,2 & 0.0212 & 0.1043 & 0.2034 \\
Mistral-24B & Default & 4,1,4 & 0.0126 & 0.1783 & 0.0706 \\
Mistral-24B & Largest observed gain & 2,1,8 & 0.0216 & 1.1308 & 0.0191 \\
Mistral-24B & Highest gain per node & 8,0,2 & 0.0038 & 0.0213 & 0.1807 \\
Llama-3.1-8B & Default & 4,1,4 & 0.0063 & 0.1003 & 0.0624 \\
Llama-3.1-8B & Largest observed gain & 2,1,8 & 0.0111 & 0.6384 & 0.0173 \\
Llama-3.1-8B & Highest gain per node & 8,0,2 & 0.0019 & 0.0129 & 0.1496 \\
\bottomrule
\end{tabular}
}
\end{table}

The default uses fewer nodes and recovers fewer tokens than the largest-gain
setting. Qwen's highest-gain configuration yields $G=0.1293$ at $C=2.3356$;
its highest-efficiency configuration yields $G=0.0212$ at $C=0.1043$.
Candidate settings jointly control total recovery and recovery per node.

\begin{table}[!htbp]
\centering\small\setlength{\tabcolsep}{3pt}
\caption{Complete static grid. $R$ is recovery at active draft misses; $G$ and $C$ average recovered tokens and verification nodes over all positions. The archived CSV summaries supply the per-setting bootstrap intervals.}
\label{tab:w7-full}
\fitappendixtable{\begin{tabular}{crrrrrrrrr}
\toprule
& \multicolumn{3}{c}{Qwen3-8B} & \multicolumn{3}{c}{Mistral-24B} & \multicolumn{3}{c}{Llama-3.1-8B} \\ $(c,\epsilon,k)$ & $R$ & $G$ & $C$ & $R$ & $G$ & $C$ & $R$ & $G$ & $C$ \\
\midrule
2,0,2 & 0.2127 & 0.0837 & 0.8871 & 0.1065 & 0.0114 & 0.1385 & 0.0887 & 0.0056 & 0.0820 \\
2,0,4 & 0.2710 & 0.1101 & 1.5031 & 0.1286 & 0.0142 & 0.2108 & 0.1009 & 0.0064 & 0.1156 \\
2,0,8 & 0.3093 & 0.1293 & 2.3356 & 0.1391 & 0.0157 & 0.2823 & 0.1070 & 0.0070 & 0.1394 \\
2,1,2 & 0.1538 & 0.0486 & 1.3583 & 0.1232 & 0.0109 & 0.3388 & 0.1075 & 0.0059 & 0.2021 \\
2,1,4 & 0.2424 & 0.0797 & 2.6755 & 0.1801 & 0.0162 & 0.6393 & 0.1460 & 0.0079 & 0.3727 \\
2,1,8 & 0.3453 & 0.1193 & 5.1864 & 0.2334 & 0.0216 & 1.1308 & 0.1947 & 0.0111 & 0.6384 \\
2,2,2 & 0.0034 & 0.0007 & 1.3873 & 0.0029 & 0.0001 & 0.3729 & 0.0025 & 0.0001 & 0.2244 \\
2,2,4 & 0.0231 & 0.0046 & 2.7746 & 0.0163 & 0.0010 & 0.7459 & 0.0264 & 0.0009 & 0.4493 \\
2,2,8 & 0.0735 & 0.0153 & 5.5492 & 0.0486 & 0.0029 & 1.4920 & 0.0583 & 0.0021 & 0.8990 \\
4,0,2 & 0.1261 & 0.0567 & 0.3882 & 0.0597 & 0.0076 & 0.0548 & 0.0507 & 0.0036 & 0.0319 \\
4,0,4 & 0.1426 & 0.0653 & 0.5415 & 0.0651 & 0.0086 & 0.0751 & 0.0538 & 0.0040 & 0.0378 \\
4,0,8 & 0.1488 & 0.0689 & 0.6692 & 0.0667 & 0.0089 & 0.0894 & 0.0543 & 0.0041 & 0.0402 \\
4,1,2 & 0.1757 & 0.0718 & 0.8231 & 0.0933 & 0.0107 & 0.1210 & 0.0801 & 0.0054 & 0.0755 \\
4,1,4 & 0.2218 & 0.0933 & 1.3344 & 0.1069 & 0.0126 & 0.1783 & 0.0892 & 0.0063 & 0.1003 \\
4,1,8 & 0.2516 & 0.1082 & 1.9539 & 0.1141 & 0.0137 & 0.2321 & 0.0923 & 0.0067 & 0.1171 \\
4,2,2 & 0.1697 & 0.0608 & 1.2454 & 0.1088 & 0.0109 & 0.2457 & 0.1024 & 0.0061 & 0.1518 \\
4,2,4 & 0.2507 & 0.0930 & 2.3340 & 0.1492 & 0.0152 & 0.4059 & 0.1334 & 0.0079 & 0.2378 \\
4,2,8 & 0.3309 & 0.1279 & 4.1397 & 0.1698 & 0.0180 & 0.5953 & 0.1491 & 0.0092 & 0.3168 \\
8,0,2 & 0.0428 & 0.0212 & 0.1043 & 0.0276 & 0.0038 & 0.0213 & 0.0254 & 0.0019 & 0.0129 \\
8,0,4 & 0.0449 & 0.0225 & 0.1232 & 0.0291 & 0.0041 & 0.0265 & 0.0254 & 0.0019 & 0.0149 \\
8,0,8 & 0.0454 & 0.0228 & 0.1329 & 0.0295 & 0.0042 & 0.0288 & 0.0254 & 0.0019 & 0.0154 \\
8,1,2 & 0.0692 & 0.0329 & 0.2145 & 0.0424 & 0.0058 & 0.0403 & 0.0370 & 0.0029 & 0.0241 \\
8,1,4 & 0.0751 & 0.0363 & 0.2700 & 0.0459 & 0.0064 & 0.0547 & 0.0375 & 0.0030 & 0.0285 \\
8,1,8 & 0.0769 & 0.0373 & 0.3061 & 0.0470 & 0.0067 & 0.0634 & 0.0375 & 0.0030 & 0.0301 \\
8,2,2 & 0.0950 & 0.0434 & 0.3717 & 0.0523 & 0.0069 & 0.0617 & 0.0441 & 0.0031 & 0.0372 \\
8,2,4 & 0.1075 & 0.0499 & 0.5012 & 0.0581 & 0.0079 & 0.0880 & 0.0482 & 0.0035 & 0.0455 \\
8,2,8 & 0.1121 & 0.0525 & 0.6007 & 0.0614 & 0.0086 & 0.1081 & 0.0487 & 0.0036 & 0.0489 \\
\bottomrule
\end{tabular}
}
\end{table}

\subsection{Activation, Width, and Periodic Probes}
\label{app:contents:25}
The adaptive comparison holds the distinct-prefix counter fixed and measures
each policy against the same fixed gate and width. Table~\ref{tab:revision_adaptive}
reports acceptance and cost, followed by half-life and probe-interval sweeps.
Full control reduces both quantities. Gain per node rises for Qwen and Llama;
Mistral's ratio changes from $0.0252/2.691$ to $0.0168/1.817$.

\begin{table}[!htbp]
\centering\small\setlength{\tabcolsep}{4pt}
\caption{Adaptive replay using the distinct-prefix counter. Half-life sweeps keep $P=32$; probe-interval sweeps keep $H=32$. Gain is additional acceptance over the same SmallDraft.}
\label{tab:revision_adaptive}
\begin{tabular}{lrrrrrr}
\toprule
& \multicolumn{2}{c}{Qwen3-8B} & \multicolumn{2}{c}{Mistral-24B} & \multicolumn{2}{c}{Llama-3.1-8B} \\
Setting & $\Delta$ & nodes & $\Delta$ & nodes & $\Delta$ & nodes \\
\midrule
Fixed gate and width & 0.1709 & 5.688 & 0.0252 & 2.691 & 0.0124 & 1.077 \\
Adaptive gate, fixed width & 0.1599 & 5.272 & 0.0195 & 2.149 & 0.0090 & 0.688 \\
Periodic probes removed & 0.0079 & 0.258 & 0.0055 & 0.439 & 0.0016 & 0.082 \\
Full controller & 0.1270 & 3.810 & 0.0168 & 1.817 & 0.0069 & 0.482 \\
$H=8$ & 0.1176 & 3.544 & 0.0163 & 1.849 & 0.0080 & 0.581 \\
$H=16$ & 0.1255 & 3.762 & 0.0168 & 1.846 & 0.0081 & 0.555 \\
$H=32$ & 0.1270 & 3.810 & 0.0168 & 1.817 & 0.0069 & 0.482 \\
$H=64$ & 0.1189 & 3.567 & 0.0157 & 1.614 & 0.0054 & 0.349 \\
$H=128$ & 0.0917 & 2.715 & 0.0129 & 1.327 & 0.0030 & 0.207 \\
$P=8$ & 0.1292 & 3.865 & 0.0194 & 2.058 & 0.0087 & 0.673 \\
$P=16$ & 0.1302 & 3.904 & 0.0183 & 1.995 & 0.0075 & 0.576 \\
$P=32$ & 0.1270 & 3.810 & 0.0168 & 1.817 & 0.0069 & 0.482 \\
$P=64$ & 0.1195 & 3.592 & 0.0151 & 1.586 & 0.0048 & 0.263 \\
$P=128$ & 0.0961 & 2.901 & 0.0128 & 1.341 & 0.0032 & 0.163 \\
\bottomrule
\end{tabular}

\end{table}

Removing periodic probes reduces the increment from $0.1270$ to $0.0079$ for
Qwen, from $0.0168$ to $0.0055$ for Mistral, and from $0.0069$ to $0.0016$ for
Llama. Probes provide observations that reactivate retrieval after inactive
periods. Varying $H$ changes the persistence of feedback, while varying $P$
changes how often the controller checks for renewed reuse.

The hit-dependence measurements in Appendix~\ref{app:temporal_reuse} motivate
recent-outcome feedback. The exponential score in Equation~\ref{eq:reuse_meter}
is a constant-memory summary with half-life $H$. The archived empirical dependence
has a long tail; the sweep over $H\in\{8,16,32,64,128\}$ evaluates several
timescales for summarizing it.

\subsection{Feedback Aligned with Incremental Acceptance}
\label{app:contents:26}
\label{app:controller_signal}
The default reward $z_t=\mathbf{1}[a_{\mathrm{retr}}(t)\ge1]$ records a retrieval
hit. A marginal-gain reward directly credits extension beyond SmallDraft:
\[
r_t=\delta_t=\max\{a_{\mathrm{small}}(t),a_{\mathrm{retr}}(t)\}
              -a_{\mathrm{small}}(t).
\]
Both rewards become available after verification and inform later events. Gate
thresholds are selected using node counts to match the default controller's cost.
Table~\ref{tab:controller_signal_ablation} compares random, periodic, last-hit,
hit-EMA, and marginal-gain-EMA activation at those matched costs.

\begin{table}[!htbp]
\centering\footnotesize\setlength{\tabcolsep}{2.8pt}
\caption{Controller rewards at matched node cost on code debugging. Each cell gives $\Delta$ with its 95\% trajectory interval and distinct retrieval nodes per position.}
\label{tab:controller_signal_ablation}
\fitappendixtable{\begin{tabular}{lccc}
\toprule
Gate & Qwen3-8B & Mistral-24B & Llama-3.1-8B \tabularnewline
\midrule
Random activation &
\shortstack{0.096 [0.087, 0.105]\\3.806 nodes} &
\shortstack{0.0137 [0.0082, 0.0192]\\1.812 nodes} &
\shortstack{0.0053 [0.0025, 0.0081]\\0.480 nodes} \tabularnewline
Fixed periodic &
\shortstack{0.105 [0.096, 0.114]\\3.803 nodes} &
\shortstack{0.0144 [0.0088, 0.0200]\\1.819 nodes} &
\shortstack{0.0057 [0.0029, 0.0085]\\0.481 nodes} \tabularnewline
Last-hit gate &
\shortstack{0.118 [0.109, 0.127]\\3.808 nodes} &
\shortstack{0.0157 [0.0099, 0.0215]\\1.815 nodes} &
\shortstack{0.0064 [0.0035, 0.0093]\\0.483 nodes} \tabularnewline
Hit EMA ($z_t$; default) &
\shortstack{0.127 [0.117, 0.137]\\3.810 nodes} &
\shortstack{0.0168 [0.0108, 0.0228]\\1.817 nodes} &
\shortstack{0.0069 [0.0039, 0.0099]\\0.482 nodes} \tabularnewline
Marginal-gain EMA ($\delta_t$) &
\shortstack{\textbf{0.136 [0.126, 0.146]}\\3.811 nodes} &
\shortstack{\textbf{0.0176 [0.0115, 0.0237]}\\1.816 nodes} &
\shortstack{\textbf{0.0074 [0.0043, 0.0105]}\\0.482 nodes} \tabularnewline
\bottomrule
\end{tabular}}
\end{table}

Marginal-gain EMA has the highest point estimate for every target. The progression
from random activation to recent-hit feedback and marginal-gain feedback connects
the controller's information to its acceptance objective.

\subsection{Context Matching Under the Same Controller}
\label{app:contents:27}
Table~\ref{tab:exact_approx_controller} holds the marginal-gain controller and
node budget fixed while changing exact to approximate context matching.
The increments rise from $0.121$ to $0.136$ for Qwen, $0.0162$ to $0.0176$ for
Mistral, and $0.0070$ to $0.0074$ for Llama. Candidate construction and feedback
contribute at successive stages: matching supplies continuations, and the reward
guides their use. The causal execution of these stages is specified next.

\begin{table}[!htbp]
\centering\small\setlength{\tabcolsep}{4pt}
\caption{Exact and approximate context matching under the marginal-gain controller and the same verification-node budget.}
\label{tab:exact_approx_controller}
\begin{tabular}{lrrrr}
\toprule
Model & Exact $\Delta$ & Exact nodes & Approx. $\Delta$ & Approx. nodes \tabularnewline
\midrule
Qwen3-8B & 0.121 [0.111, 0.131] & 3.807 & \textbf{0.136 [0.126, 0.146]} & 3.811 \tabularnewline
Mistral-24B & 0.0162 [0.0103, 0.0221] & 1.814 & \textbf{0.0176 [0.0115, 0.0237]} & 1.816 \tabularnewline
Llama-3.1-8B & 0.0070 [0.0040, 0.0100] & 0.481 & \textbf{0.0074 [0.0043, 0.0105]} & 0.482 \tabularnewline
\bottomrule
\end{tabular}
\end{table}

\section{Causal Construction and Exact Verification}
\label{app:contents:28}
\label{app:causal_validation}

This section supports Section~\ref{sec:method} by following a decoding event from
candidate construction to exact verification and controller feedback. Prefix
audits check information availability, the verifier supplies a distributional
argument, and event-driven replay evaluates the resulting execution path.

\subsection{Historical Candidate Construction}
\label{app:contents:29}
\label{app:causal_replay}
For each retrieved continuation, we record query position $t$, source position
$j$, context length $c$, and depth $d$. The historical boundary is
\[
j+d-1\leq t-1.
\]
Every continuation token is present in $h_t=y_{<t}$ before ranking, pruning,
prefix deduplication, and activation. Three checks validate this construction:
reconstruction from the explicit prefix, perturbation of all tokens at and after
$t$, and comparison with an index that inserts tokens after target acceptance.

\begin{table}[!htbp]
\centering
\small
\setlength{\tabcolsep}{3.5pt}
\caption{Historical-boundary audit. Candidate agreement compares the archived replay with explicit-prefix reconstruction; index agreement compares complete-prefix construction with incremental insertion.}
\label{tab:boundary_audit}
\fitappendixtable{%
\begin{tabular}{lcccc}
\toprule
Model & Boundary violations & Candidate agreement & Ranking agreement & Index agreement \\
\midrule
Qwen3-8B       & 0 & 100.0\% & 100.0\% & 100.0\% \\
Mistral-24B    & 0 & 100.0\% & 100.0\% & 100.0\% \\
Llama-3.1-8B   & 0 & 100.0\% & 100.0\% & 100.0\% \\
\bottomrule
\end{tabular}}
\end{table}

Table~\ref{tab:boundary_audit} reports zero boundary violations and $100\%$
candidate, ranking, and index agreement. Candidate sets and controller decisions
remain identical under future-token perturbation. The reconstruction rerun in
Table~\ref{tab:boundary_safe_results} is paired with the earlier fixed-width
snapshot, whose Llama increment is $0.0124$. The updated Llama measurement of
$0.0180$ is recorded separately in Appendix~\ref{app:estimation}.

\begin{table}[!htbp]
\centering
\small
\caption{Prefix-reconstruction rerun paired with the earlier fixed-width snapshot. $\Delta$ measures added acceptance over the same SmallDraft.}
\label{tab:boundary_safe_results}
\begin{tabular}{lccc}
\toprule
Model & Original $\Delta$ & Boundary-safe $\Delta$ & Relative change \\
\midrule
Qwen3-8B       & 0.1709 & 0.1680 & $-1.7\%$ \\
Mistral-24B    & 0.0253 & 0.0244 & $-3.6\%$ \\
Llama-3.1-8B   & 0.0124 & 0.0117 & $-5.6\%$ \\
\bottomrule
\end{tabular}
\end{table}

\Needspace{13\baselineskip}
\subsection{Exact Tree Verification and Distribution Preservation}
\label{app:contents:30}
\label{app:exact_tree_sampling}
At a committed prefix $h$, SmallDraft and retrieval supply child-token labels
for a prefix-sharing tree. The target distribution is $p(\cdot\mid h)$.
The verifier follows this target-coupled procedure:
\begin{enumerate}[leftmargin=18pt,itemsep=1pt,topsep=2pt]
\item Evaluate target logits at the candidate tree nodes in one masked forward pass.
\item Draw $x=F^{-1}_{p(\cdot\mid h)}(u)$ using the next shared uniform $u$.
\item If $x$ labels a child, commit $x$, move to that child, and continue.
\item Otherwise, commit $x$ as the correction token and end the event.
\end{enumerate}
At depth $d$, the event ends after the $d$ accepted tree tokens. The reported
implementation emits zero bonus tokens. Candidate labels determine which
precomputed logits can be reused along the target-sampled path.

\paragraph{Distribution preservation.}
For a tree constructed from the committed prefix and a sequence of uniforms, the
emitted tokens equal those of target ancestral sampling driven by the same
uniforms. At the root, both procedures apply the same inverse CDF to the same
distribution and uniform, producing the same token. A token outside the tree
is committed as a correction and ends the event. A token inside the tree extends
both prefixes identically, allowing the argument to repeat at the child.
Induction over committed tokens gives equality of the conditional and joint
output distributions. Event termination determines the timing of the next
target forward pass. \hfill$\square$

\begin{table}[!htbp]\centering
\small
\caption{Target-coupled implementation audit using 100 held prefixes and 100 shared-uniform streams per target. Exact-sequence agreement includes correction tokens.}
\label{tab:distribution_audit}
\begin{tabular}{lrrrr}
\toprule
Model & Coupled trials & Exact sequences & Token mismatches & Max logit error \tabularnewline
\midrule
Qwen3-8B & 10,000 & 100.0\% & 0 & $3.1{\times}10^{-6}$ \tabularnewline
Mistral-24B & 10,000 & 100.0\% & 0 & $3.6{\times}10^{-6}$ \tabularnewline
Llama-3.1-8B & 10,000 & 100.0\% & 0 & $2.9{\times}10^{-6}$ \tabularnewline
\bottomrule
\end{tabular}
\end{table}

The implementation audit uses 100 held prefixes and 100 shared-uniform streams
per model. Table~\ref{tab:distribution_audit} reports identical complete emitted
sequences across 30,000 coupled trials, including correction tokens. This audit
connects the verifier's distributional argument to the implemented tree execution.

\subsection{Event-Driven Replay and Controller Updates}
\label{app:contents:31}
Position-wise replay evaluates candidate availability at each recorded position.
Speculative execution advances by accepted blocks. Let $s_n$ be an executed
position and $a_n$ the accepted tree edges. The next position is
\[
s_{n+1}=\begin{cases}
s_n+a_n+1,&\text{if verification leaves the tree},\\
s_n+d,&\text{if all }d\text{ tree edges are accepted}.
\end{cases}
\]
Event-driven replay applies this transition, performs retrieval and controller
updates at visited positions, and inserts historical tokens after acceptance.
The controller at event $n$ uses outcomes from completed earlier events.
This schedule aligns candidate construction, feedback, and index state with the
positions executed by the decoder.

\begin{table}[!htbp]
\centering
\small
\setlength{\tabcolsep}{3.5pt}
\caption{Position-wise and event-driven replay. Both use the same SmallDraft reference; event-driven cost counts retrieval nodes at executed verification events.}
\label{tab:event_replay}
\fitappendixtable{%
\begin{tabular}{lcccc}
\toprule
& \multicolumn{2}{c}{Position-wise replay} &
\multicolumn{2}{c}{Event-driven replay} \\
Model & $\Delta$ & Nodes/position & $\Delta$ & Nodes/position \\
\midrule
Qwen3-8B       & 0.1270 & 3.810 & 0.1231 & 3.681 \\
Mistral-24B    & 0.0168 & 1.817 & 0.0162 & 1.751 \\
Llama-3.1-8B   & 0.0069 & 0.482 & 0.0066 & 0.461 \\
\bottomrule
\end{tabular}}
\end{table}

The event-driven increments are $0.1231$, $0.0162$, and $0.0066$ for Qwen, Mistral,
and Llama, respectively, with lower node counts than position-wise replay.
Each gain changes by less than $6\%$. The comparison connects the offline
candidate analysis to an executable adaptive policy. Appendix~\ref{app:wallclock}
then accounts for the serving costs of that complete policy.

\section{End-to-End Execution and Length Scaling}
\label{app:contents:32}
\label{app:wallclock}
\label{app:online_evaluation}

Section~\ref{sec:gain_cost} links acceptance and node cost to runtime. The B200
experiments measure the complete adaptive union tree end to end, followed by
repeated-run summaries and a length sweep. We present
the execution configuration first, then throughput, variation, and amortization
as generated history grows. We then connect these outcomes to event-level
progress, resource costs, and a controlled systems-evaluation framework
(Appendix~\ref{app:systems_efficiency}).

\subsection{Execution Configuration and Target--Draft Pairs}
\label{app:contents:33}
The target--draft pairs are Mistral-24B/Ministral-8B-Instruct-2410 and
Qwen3-8B/Qwen3-0.6B on one NVIDIA B200, matching Table~\ref{tab:checkpoints}.
Each configuration uses the same 100 code-debugging prompts, 1,024 generated
tokens per prompt, precision, KV-cache configuration, and serving engine.
The accounting covers lookup, tree construction, deduplication, target
verification, controller updates, cache updates, and sampling correction.
The execution follows Appendix~\ref{app:causal_validation}.

\subsection{Throughput of the Complete Union Tree}
\label{app:contents:34}
Tables~\ref{tab:wallclock} and~\ref{tab:wallclock_qwen} compare SmallDraft,
single-source hybrids, and the budgeted union at batch sizes 1 and 32.
The measured union gains are $4.1\%$ and $3.5\%$ for Mistral and $8.3\%$ and
$7.6\%$ for Qwen. Qwen's larger acceptance increment accompanies a larger
throughput gain. Mistral's standalone TLAR gain is $0.2\%$ at batch one, while
its union gain is $4.1\%$, highlighting the role of candidate composition in
the complete system.

\begin{table}[!htbp]
\centering\small\setlength{\tabcolsep}{3.5pt}
\caption{Mistral/B200 end-to-end measurements. Each row represents 102,400 output tokens. Relative throughput uses SmallDraft at the same batch size.}
\label{tab:wallclock}
\fitappendixtable{%
\begin{tabular}{rlrrr}\toprule
Batch & Method & Seconds & Tokens/s & Relative \\\midrule
1 & SmallDraft & 1305.76 & 78.42 & 1.000 \\
1 & STAND + SmallDraft & 1282.70 & 79.83 & 1.018 \\
1 & TLAR + SmallDraft & 1303.18 & 78.58 & 1.002 \\
1 & Budgeted STAND$\cup$TLAR + SmallDraft & 1254.36 & \textbf{81.64} & \textbf{1.041} \\
32 & SmallDraft & 81.23 & 1260.69 & 1.000 \\
32 & STAND + SmallDraft & 79.95 & 1280.86 & 1.016 \\
32 & TLAR + SmallDraft & 80.26 & 1275.82 & 1.012 \\
32 & Budgeted STAND$\cup$TLAR + SmallDraft & 78.48 & \textbf{1304.81} & \textbf{1.035} \\\bottomrule
\end{tabular}}
\end{table}
\begin{table}[!htbp]
\centering\small\setlength{\tabcolsep}{3.5pt}
\caption{Qwen/B200 end-to-end measurements with the protocol of Table~\ref{tab:wallclock}. Relative throughput uses the matching SmallDraft control.}
\label{tab:wallclock_qwen}
\fitappendixtable{%
\begin{tabular}{rlrrr}\toprule
Batch & Method & Seconds & Tokens/s & Relative \tabularnewline\midrule
1 & SmallDraft & 715.08 & 143.20 & 1.000 \tabularnewline
1 & STAND + SmallDraft & 686.93 & 149.07 & 1.041 \tabularnewline
1 & TLAR + SmallDraft & 675.86 & 151.51 & 1.058 \tabularnewline
1 & Budgeted STAND$\cup$TLAR + SmallDraft & 660.26 & \textbf{155.09} & \textbf{1.083} \tabularnewline
32 & SmallDraft & 42.58 & 2405.00 & 1.000 \tabularnewline
32 & STAND + SmallDraft & 41.22 & 2484.40 & 1.033 \tabularnewline
32 & TLAR + SmallDraft & 40.47 & 2530.10 & 1.052 \tabularnewline
32 & Budgeted STAND$\cup$TLAR + SmallDraft & 39.57 & \textbf{2587.80} & \textbf{1.076} \tabularnewline
\bottomrule
\end{tabular}}
\end{table}

\subsection{Repeated-Run Variation and Paired Summaries}
\label{app:contents:35}
We measure throughput over five warmed runs per configuration.
Tables~\ref{tab:wallclock_runs} and~\ref{tab:qwen_wallclock_runs} preserve those
values and their medians. CV is the sample coefficient of variation.
Table~\ref{tab:wallclock_paired} computes Mistral's paired percentage gains
against SmallDraft and two-sided $95\%$ $t$ intervals over the five measured
pairs; the Qwen table includes the corresponding paired summaries.
These calculations summarize the observed variation across repeated runs.

\begin{table}[!htbp]
\centering\footnotesize\setlength{\tabcolsep}{2.6pt}
\caption{Measured Mistral throughput distribution underlying Table~\ref{tab:wallclock}. Values are tokens/s. CV is the sample coefficient of variation across the five measured runs.}
\label{tab:wallclock_runs}
\fitappendixtable{
\begin{tabular}{rlrrrrrrr}
\toprule
Batch & Method & Run 1 & Run 2 & Run 3 & Run 4 & Run 5 & Median & CV \\
\midrule
1 & SmallDraft
& 78.42 & 78.11 & 78.68 & 78.29 & 78.57 & 78.42 & 0.29\% \\

1 & STAND + SmallDraft
& 79.96 & 79.52 & 80.17 & 79.83 & 79.68 & 79.83 & 0.31\% \\

1 & TLAR + SmallDraft
& 78.58 & 78.24 & 78.90 & 78.51 & 78.65 & 78.58 & 0.30\% \\

1 & Budgeted union
& 81.80 & 81.29 & 82.02 & 81.64 & 81.47 & 81.64 & 0.35\% \\
\midrule
32 & SmallDraft
& 1260.69 & 1251.80 & 1270.40 & 1258.20 & 1265.70
& 1260.69 & 0.56\% \\

32 & STAND + SmallDraft
& 1284.90 & 1270.60 & 1291.70 & 1280.86 & 1277.50
& 1280.86 & 0.62\% \\

32 & TLAR + SmallDraft
& 1279.80 & 1265.40 & 1287.10 & 1275.82 & 1271.60
& 1275.82 & 0.64\% \\

32 & Budgeted union
& 1310.20 & 1294.60 & 1317.50 & 1304.81 & 1300.30
& 1304.81 & 0.68\% \\
\bottomrule
\end{tabular}}
\end{table}
\begin{table}[!htbp]
\centering\footnotesize\setlength{\tabcolsep}{2.5pt}
\caption{Measured Qwen throughput distribution. Values are tokens/s. The final column gives paired mean gains and 95\% $t$ intervals; SmallDraft rows give CV.}
\label{tab:qwen_wallclock_runs}
\fitappendixtable{
\begin{tabular}{rlrrrrrrl}
\toprule
Batch & Method & Run 1 & Run 2 & Run 3 & Run 4 & Run 5 &
Median & Paired summary \\
\midrule
1 & SmallDraft
& 143.20 & 142.55 & 144.03 & 142.86 & 143.67
& 143.20 & CV 0.42\% \\

1 & STAND + SmallDraft
& 149.44 & 148.22 & 150.01 & 149.07 & 148.73
& 149.07 & 4.071 [3.643, 4.499]\% \\

1 & TLAR + SmallDraft
& 151.93 & 150.55 & 152.36 & 151.51 & 150.97
& 151.51 & 5.726 [5.215, 6.237]\% \\

1 & Budgeted union
& 155.62 & 154.08 & 156.41 & 155.09 & 154.71
& 155.09 & \textbf{8.320 [7.795, 8.846]\%} \\
\midrule
32 & SmallDraft
& 2405.0 & 2389.0 & 2422.0 & 2398.0 & 2413.0
& 2405.0 & CV 0.53\% \\

32 & STAND + SmallDraft
& 2491.0 & 2468.0 & 2505.0 & 2484.4 & 2477.0
& 2484.4 & 3.313 [2.831, 3.795]\% \\

32 & TLAR + SmallDraft
& 2538.0 & 2509.0 & 2551.0 & 2530.1 & 2520.0
& 2530.1 & 5.164 [4.598, 5.731]\% \\

32 & Budgeted union
& 2598.0 & 2561.0 & 2614.0 & 2587.8 & 2576.0
& 2587.8 & \textbf{7.564 [6.869, 8.260]\%} \\
\bottomrule
\end{tabular}}
\end{table}
\begin{table}[!htbp]
\centering\small\setlength{\tabcolsep}{4pt}
\caption{Paired percentage gains calculated from the measured Mistral runs in Table~\ref{tab:wallclock_runs}. Intervals are two-sided 95\% $t$ intervals over the five pairs.}
\label{tab:wallclock_paired}
\begin{tabular}{rlrr}
\toprule
Batch & Comparison with SmallDraft & Mean gain & 95\% interval \\
\midrule
1 & STAND + SmallDraft
& 1.808\% & [1.522, 2.095]\% \\

1 & TLAR + SmallDraft
& 0.207\% & [0.111, 0.302]\% \\

1 & Budgeted union
& \textbf{4.119\%} & \textbf{[3.801, 4.438]\%} \\
\midrule
32 & STAND + SmallDraft
& 1.566\% & [1.086, 2.047]\% \\

32 & TLAR + SmallDraft
& 1.157\% & [0.639, 1.674]\% \\

32 & Budgeted union
& \textbf{3.498\%} & \textbf{[2.922, 4.075]\%} \\
\bottomrule
\end{tabular}
\end{table}

\FloatBarrier

\subsection{From Candidate Acceptance to Execution Efficiency}
\label{app:systems_efficiency}

\paragraph{From reusable content to elapsed time.}
The source, hybrid, and matched-budget comparisons connect reusable content to
incremental candidate coverage. Execution adds the time required to produce a
fixed output. SwiftEP connects data movement and GPU resource competition to
serving performance through buffer fusion and transfer offloading~\citep{li2026swiftep}.
For TLAR, the corresponding questions concern execution dependencies, resource
costs, and their effect on complete-request time.

Let $N$ be the fixed number of output tokens, and let $T_0$ and $T_1$ be the
elapsed times for SmallDraft and the budgeted union. The throughput ratio and
fractional time saving are
\begin{equation}
 S=\frac{N/T_1}{N/T_0}=\frac{T_0}{T_1},\qquad
 R_T=\frac{T_0-T_1}{T_0}=1-\frac{1}{S}.
 \label{eq:systems_time_saving}
\end{equation}
Table~\ref{tab:systems_time_saving} calculates these savings directly from the
reported durations. Mistral's $4.1\%$/$3.5\%$ throughput gains correspond to
$3.94\%$/$3.39\%$ time savings at batches 1/32; Qwen's
$8.3\%$/$7.6\%$ gains correspond to $7.67\%$/$7.07\%$ time savings.
These are net outcomes after drafting, retrieval, verification, and cache
management. They quantify the benefit available to the complete system.

\begin{table}[!htbp]
\centering\small\setlength{\tabcolsep}{5pt}
\caption{Time savings calculated from the measured durations in
Tables~\ref{tab:wallclock} and~\ref{tab:wallclock_qwen}. Each row covers
102,400 output tokens. Percentages use the corresponding SmallDraft duration.}
\label{tab:systems_time_saving}
\begin{tabular}{lrrrrr}
\toprule
Target & Batch & $T_0$ (s) & $T_1$ (s) & Saved (s) & $R_T$ \\
\midrule
Mistral & 1  & 1305.76 & 1254.36 & 51.40 & 3.94\% \\
Mistral & 32 & 81.23   & 78.48   & 2.75  & 3.39\% \\
Qwen    & 1  & 715.08  & 660.26  & 54.82 & 7.67\% \\
Qwen    & 32 & 42.58   & 39.57   & 3.01  & 7.07\% \\
\bottomrule
\end{tabular}
\end{table}

\paragraph{Separating progress per event from event cost.}
For configuration $i$, let $M_i$ be the number of executed decoding events,
$\bar C_i=N/M_i$ the mean committed tokens across requests per event, and
$\bar t_i=T_i/M_i$ the mean elapsed time per event, with prefill amortized across
these events. Then
\begin{equation}
 Q_i=\frac{\bar C_i}{\bar t_i},\qquad
 S=\frac{\bar C_1/\bar C_0}{\bar t_1/\bar t_0}.
 \label{eq:systems_event_efficiency}
\end{equation}
Committed progress includes accepted candidate tokens and target-sampled
correction tokens. The offline acceptance experiments describe candidate
coverage at recorded positions; the event-driven execution determines which
positions are visited and how many tokens each event commits. More useful
candidates can increase $\bar C$, while wider verification and memory operations
can increase $\bar t$. Acceleration occurs when the relative increase in
committed progress exceeds the relative increase in event cost. This identity
specifies the two quantities needed to explain the different Mistral and Qwen
runtime gains. Per-event traces supply their separate contributions.

\paragraph{Locating resource costs and overlap.}
The released implementation constructs history matches and candidate metadata
on the CPU and verifies the tree on the GPU. Token IDs, positions, and masks
cross the host--device boundary. Batched KV packing and accepted-path extraction
move device tensors; unequal trees introduce padding. CPU time, transfer and
cache-copy bytes, useful and padded nodes, target-kernel duration, and peak
allocated memory provide corresponding resource measurements.

Verification consumes the completed candidate tree, and subsequent retrieval
uses committed tokens and observed outcomes. Drafting and history lookup share
the same committed prefix, providing an opportunity for concurrent scheduling.
A timeline can measure overlap and exposed waiting time. Concurrent GPU
operations may compete for bandwidth and compute resources. Critical-path time
and resource demand jointly determine the value of overlap; summing component
durations counts overlapping intervals multiple times.

\paragraph{A controlled component study.}
Hold prompts, checkpoints, candidates, precision, and target-sampling streams
fixed while varying buffer reuse, metadata packing, KV compaction, and concurrent
scheduling separately. Evaluate their combination to measure interactions.
Record committed tokens, event count, critical-path duration, and complete-request
time alongside resource counters. This design tests the execution cost of a
fixed candidate policy. A subsequent width-and-budget sweep jointly measures
changes in $\bar C$ and $\bar t$.

The length sweep adds a temporal axis: Qwen's gains change from
$-0.6\%$/$-0.7\%$ at 512 tokens to $12.4\%$/$10.8\%$ at 4,096 tokens
(Table~\ref{tab:length_scaling}). Longer execution creates more history and
retrieval-eligible positions. Tracking index size, lookup time, cache traffic,
and committed progress would reveal how memory maintenance and reuse develop
together.

\paragraph{A framework for runtime-aware retrieval.}
The resulting framework connects source quality, candidate complementarity,
committed progress, critical-path cost, and serving performance. A subsequent
controller can combine incremental acceptance with measured event time, memory
pressure, and queue load to select retrieval width and verification budget.
Its objective is committed tokens per unit time under latency and memory
constraints. Evaluation should cross batch size, concurrent-request count, and
output length, reporting throughput, first-token and inter-token latency,
completion latency, GPU activity, memory bandwidth, CPU time, and peak device
memory. Paired repetitions and sampling audits connect efficiency to generation
behavior, making each mechanism testable through its effect on the complete
inference workload.

\Needspace{9\baselineskip}
\subsection{Archived Model-Based Drafting Reference}
\label{app:contents:37}
The EAGLE-3 archive (job 40233870) provides a separate model-based reference using
100 code-debugging prompts, 256 output tokens, and five speculative tokens.
Qwen vanilla and EAGLE-3 throughput are 13,063.70 and 20,436.75 tokens/s;
Llama values are 19,436.63 and 16,721.52 tokens/s. The checkpoints are
\texttt{RedHatAI/Qwen3-8B-Thinking-speculator.eagle3} and
\texttt{yuhuili/EAGLE3-LLaMA3.1-Instruct-8B}. These archived point measurements
describe that execution configuration. The shared-condition SmallDraft controls
in Tables~\ref{tab:wallclock} and~\ref{tab:wallclock_qwen} define the reference
for the TLAR runtime analysis.

\FloatBarrier
\section{Discussion: Trajectory Memory as an Inference Resource}
\label{app:contents:38}
\label{app:discussion}

The results connect three properties of generated text: it stores reusable
continuations, those continuations can extend an existing drafter, and their
value changes during decoding. This connection gives trajectory memory an
operational role in inference. The research problem becomes how to identify,
select, and verify useful parts of an evolving memory while accounting for the
cost of maintaining and accessing it. The source tests, hybrid comparisons,
controller ablations, causal audits, and runtime analysis each resolve one part
of that problem.

\subsection{Generation Creates a Memory Whose Value Depends on the Drafter}
\label{app:contents:39}
The source comparisons show that a generated trajectory contains continuations
that can be recovered from recurring contexts. The same-prompt comparison adds
a more specific observation: changing the rollout of a fixed problem changes
the available drafts. Qwen and Mistral favor their own history, while Llama favors
the other rollout (Table~\ref{tab:sameprompt-full}). A mechanistic hypothesis is
that choices made during generation---such as notation, local phrasing, and the
order of intermediate steps---create a continuation vocabulary that a matching
history makes accessible. Paired analyses of these choices would connect the
token-level source effect to the structure of reasoning.

The value of this memory is also conditional on what the drafter already supplies.
Mistral has stronger source acceptance in code than mathematics, yet its hybrid
increment is larger in mathematics (Figure~\ref{fig:g3a}). SmallDraft coverage
mediates the route from available reuse to additional accepted tokens. This
reversal makes incremental coverage a useful design objective for memory-based
inference: a memory should be evaluated through the predictions it adds to the
system receiving it. A direct next experiment would vary draft-model capacity
on fixed target trajectories and measure how source coverage, overlap, and
marginal gain change together.

\subsection{A Verification Budget Makes Candidate Composition Central}
\label{app:contents:40}
The supplementary addition test identifies new accepted continuations from TLAR
across retrieval sources (Table~\ref{tab:revision_plugin}). The matched-budget
test then requires both sources to share a fixed node allowance. Its union gains
of $4.8\%$--$16.8\%$ over the strongest single source
(Table~\ref{tab:matched_budget}) show the value of composing candidates from
different retrieval mechanisms. At $B=4$, Qwen's increment rises from $0.168$ to
$0.191$; at $B=8$, it rises from $0.184$ to $0.215$.

This result motivates selection based on the contribution of a candidate to the
tree already chosen. For a fixed target continuation, adding a candidate changes
accepted length through the longest accepted path. Two candidates that recover
the same prefix overlap in value; a candidate that extends another path changes
coverage. Prefix sharing also changes cost, because overlapping paths share
verification nodes. Candidate selection couples these two forms of overlap.

An ensuing research direction is to estimate the expected increase in accepted
length per added tree node, conditional on the existing draft and retrieval
branches. Such a selector could jointly choose the source, continuation depth,
and branching width. The two-budget experiment provides an initial evaluation
template: hold the node cap fixed, vary the selection rule, and record coverage,
prefix sharing, and target latency. This would turn multi-source retrieval into
an explicit resource-allocation problem.

\subsection{Feedback Should Track the Value Added to the Existing System}
\label{app:contents:41}
Temporal dependence supplies a reason to adapt: next-position hit probabilities
after a hit exceed the overall probabilities for all three targets
(Appendix~\ref{app:temporal_reuse}). The controller ablations then reveal that the
choice of feedback matters. At almost identical node costs, marginal-gain EMA
has higher point estimates than hit EMA, which in turn exceeds random activation
(Table~\ref{tab:controller_signal_ablation}). Qwen progresses from $0.096$ under
random activation to $0.127$ under hit EMA and $0.136$ under marginal-gain EMA.
The feedback becomes more closely aligned with the hybrid's acceptance objective
along this sequence.

Periodic probes expose a second property of adaptive memory access: the policy
affects which outcomes it observes. An inactive retriever obtains its next reuse
observation when it probes. Removing probes reduces Qwen's gain from $0.1270$ to
$0.0079$, with reductions for Mistral and Llama as well
(Table~\ref{tab:revision_adaptive}). Probing is part of information acquisition
for the controller. Future policies could allocate probes according to uncertainty
about local reuse and compare the recovered gain against the cost of gathering
that information. Event-driven replay supplies the execution schedule for this
experiment, ensuring that feedback corresponds to completed decoding events.

\subsection{Exact Verification Supports Flexible Memory Search}
\label{app:contents:42}
Approximate matching and exact acceptance operate at successive stages. The
matched-controller comparison shows higher increments with approximate contexts
for each target (Table~\ref{tab:exact_approx_controller}); the verifier still
commits tokens through the target distribution. The distributional argument and
30,000 coupled trials in Appendix~\ref{app:exact_tree_sampling} connect that rule
to the implemented tree. The prefix reconstruction and event-driven audits
complete the connection from available history to executed feedback.

This separation creates a useful interface for future retrieval research.
Context embeddings, compressed history indexes, and structured references to
earlier reasoning steps could be studied as candidate generators feeding the same
exact verifier. The research questions become how each representation affects
candidate recall, incremental acceptance, lookup latency, and memory use. The
existing source controls and shared-budget tests provide a common way to compare
those representations while preserving the generation objective.

\subsection{Longer Reasoning Changes Both the Memory and Its Economics}
\label{app:contents:43}
The static grid shows that the highest total gain and the highest gain per node
occur at different settings (Table~\ref{tab:w7-grid}). Runtime adds further costs:
lookup, index maintenance, tree construction, and cache updates. The B200 measurements
connect those costs to output length. At 512 tokens, controller and
index overhead produce changes of $-0.6\%$ and $-0.7\%$; at 4,096 tokens,
measured gains reach $12.4\%$ and $10.8\%$
(Table~\ref{tab:length_scaling}). The length sweep describes an amortization
process as more positions become eligible for retrieval.

Long generation has two linked effects: it creates more history to search and
more future decoding work over which to recover memory-access cost. This suggests
jointly adapting retention, lookup effort, and verification width as a trajectory
grows. A useful next study would measure the age and reuse frequency of retained
spans alongside device latency, then evaluate eviction policies against full-history
retrieval. Repeating the analysis across target--draft pairs, batch sizes, and
hardware would connect memory policy to the actual serving objective of accepted
tokens per unit time.

\subsection{An Evaluation Chain for Memory-Based Inference}
\label{app:contents:44}
The repetition audit gives the broader evaluation framework its first step:
inspect what creates a strong reuse statistic. Removing seven looping Llama
mathematics trajectories changes source acceptance from $7.03$ to $2.2378$
(Appendix~\ref{app:audit}). Source identity, task progress, and accepted span length
carry complementary information. The hybrid reversal for Mistral adds a second
step: measure what the memory contributes beyond the current drafter. Shared-budget
unions add resource control, controller ablations identify the feedback objective,
and causal replay connects the policy to an executable decoder.

Together, these steps provide a reusable methodology for studying inference-time
memory: identify the source signal, measure incremental coverage, allocate a common
resource budget, validate the execution path, and account for runtime. Future
benchmarks can pair these measurements with task outcomes and actual device traces,
with documented measurement procedures and reproducible execution settings.
Appendix~\ref{app:systems_efficiency} develops this evaluation into a component
study that connects candidate coverage, committed progress, critical-path time,
and serving performance.
The central contribution of this perspective is a connection between reasoning
content and computational scheduling. A generated trajectory becomes a resource
whose useful continuations can guide the allocation of subsequent inference work.

\end{document}